\pdfoutput=1

\documentclass[11pt]{article}

\usepackage{acl}

\usepackage{times}
\usepackage{latexsym}
\usepackage{booktabs}
\usepackage{graphicx}
\usepackage{amsmath}
\usepackage{enumitem}
\usepackage{float}
\usepackage{multirow}
\usepackage{color, colortbl}
\definecolor{Gray}{gray}{0.9}

\newcommand{\framename}{\textsc{SNaC}}
\usepackage{url}
\usepackage{dingbat}

\usepackage[textsize=scriptsize]{todonotes}

\usepackage[T1]{fontenc}

\usepackage[utf8]{inputenc}

\usepackage{microtype}

\usepackage{inconsolata}
\title{\framename: Coherence Error Detection for Narrative Summarization}

\author{Tanya Goyal$^1$ \hspace{0.7cm} Junyi Jessy Li$^2$ \hspace{0.7cm} Greg Durrett$^1$ \\
    $^1$ Department of Computer Science \hspace{0.7cm}
    $^2$ Department of Linguistics \\
  The University of Texas at Austin \\
  {\tt tanyagoyal@utexas.edu}}

\begin{document}
\maketitle
\begin{abstract}
Progress in summarizing long texts is inhibited by the lack of appropriate evaluation frameworks. A long summary that appropriately covers the facets of that text must also present a coherent narrative, but current automatic and human evaluation methods fail to identify gaps in coherence. In this work, we introduce \framename, a narrative coherence evaluation framework for fine-grained annotations of long summaries. We develop a taxonomy of coherence errors in generated narrative summaries and collect span-level annotations for 6.6k sentences across 150 book and movie summaries. Our work provides the first characterization of coherence errors generated by state-of-the-art summarization models and a protocol for eliciting coherence judgments from crowdworkers. Furthermore, we show that the collected annotations allow us to benchmark past work in coherence modeling and train a strong classifier for automatically localizing coherence errors in generated summaries. Finally, our \framename~framework can support future work in long document summarization and coherence evaluation, including improved summarization modeling and post-hoc summary correction.\footnote{All collected annotations and models released at: \url{https://github.com/tagoyal/snac}.}
\end{abstract}

\section{Introduction}
As pre-trained models for news summarization  \cite{lewis2020bart, zhang2020pegasus, brown2020language} have improved drastically, researchers have begun tackling increasingly challenging settings, particularly long document summarization and generation of longer summaries \cite{kryscinski2021booksum, huang2021efficient, zhang2021summ, wu2021recursively}. 
Summaries in these settings 
differ considerably from the newswire summaries of past research efforts \cite{nallapati2016abstractive, narayan2018don}: models now need to extract salient information from different parts of a significantly longer document, and na\"{i}vely combining these in a much longer output is less likely to yield a summary with coherent discourse structure. 

\begin{figure}[t]
    \centering
    \includegraphics[trim=60mm 133mm 15mm 48mm,scale=0.28, clip]{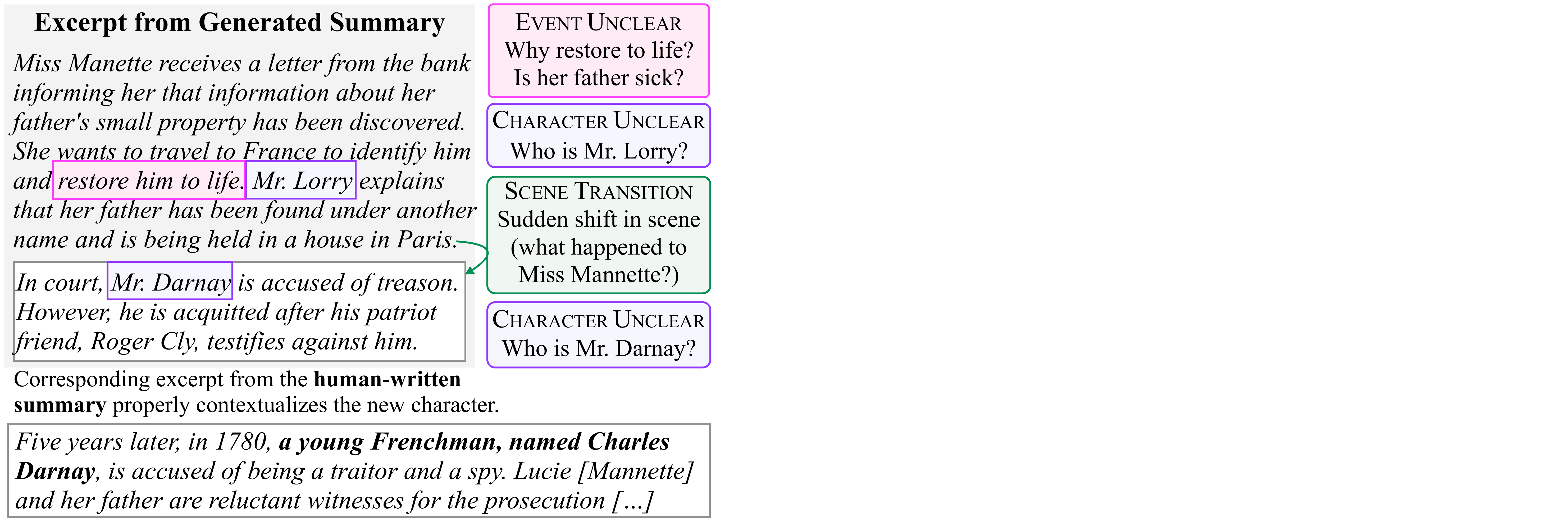}
    \caption{Excerpt from a generated book summary by OpenAI's 175B model \cite{wu2021recursively}. 
    Individual segments do not follow a coherent structure and extra information is often needed to understand the narrative.} 
    \label{fig:intro}
\end{figure}

\begin{figure*}[t]
    \centering
    \includegraphics[trim=25mm 220mm 40mm 10mm,scale=0.26, clip]{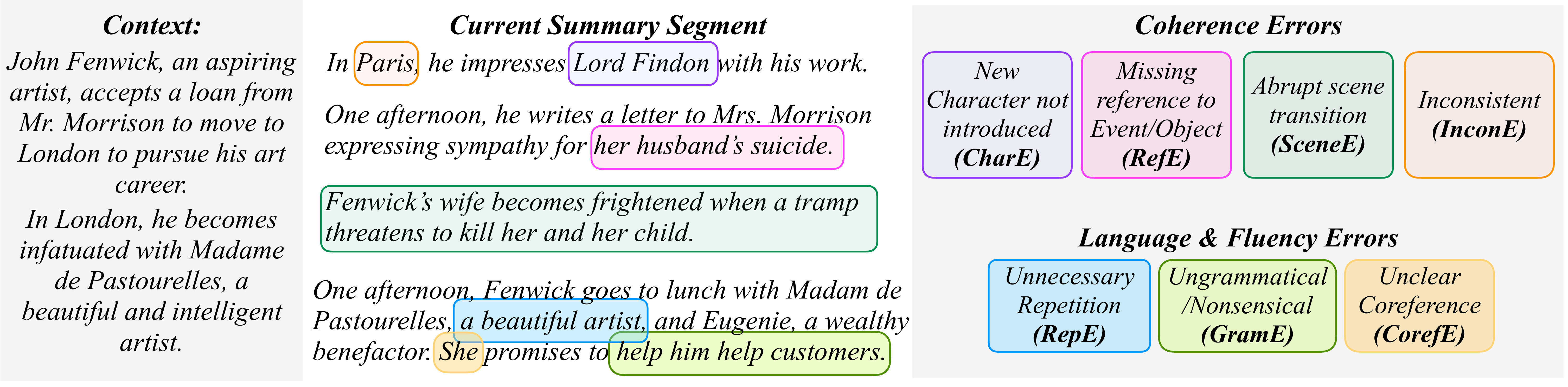}
    \caption{\framename's error schema. Given context, i.e., the generated summary until that point, annotators identify error spans in the current summary segment. 
    We define two high-level error categories: (1) Coherence Errors that directly affect narrative understanding, and (2) Language Errors that measure other aspects, like grammar.} 
    \label{fig:error_types}
\end{figure*}

This shift in the scope of the summarization task calls for a reexamination of the summarization evaluation framework. Even for short newswire summaries, \citet{fabbri2021summeval} showed that automated metrics are inadequate, and consequently, reporting results from a human evaluation study has become the standard practice. However, human evaluation is rarely done for longer summaries possibly due to the associated labor costs of reading and evaluating long text.
It is also unclear whether A/B testing or Likert-scale based annotation frameworks transfer to long summary settings. Establishing human evaluation protocols is critical for comparing different modeling approaches and measuring progress. 

Recently, \citet{wu2021recursively} proposed a strong book summarization model but showed that although generated summaries covered \textit{important} information from the books, they read like a list of events stapled together without any coherent narrative structure (see Figure \ref{fig:intro}). We found similar characteristics in other recent narrative summarization models \cite{kryscinski2021booksum,zhang2021summ}. Now that models are so good at generating fluent and on-topic sentences, the \emph{coherence of the whole summary} becomes a first-order issue that must be evaluated in these new settings.


In this work, we introduce \framename, a framework for collecting fine-grained annotations to evaluate \textbf{S}ummary \textbf{Na}rrative \textbf{C}oherence. We develop an error schema with 7 narrative error types grounded in actual errors made by current summarization models. Our fine-grained taxonomy allows annotators to explicitly state what kind of coherence error exists in a summary and pinpoint where it occurs. We show that such a fine-grained annotation framework is better suited for collecting crowd annotations than a Likert scale-based holistic evaluation of coherence. 

We enlist crowdworkers to collect a large-scale dataset of 9.6k span-level error annotations in narrative summaries generated by current state-of-the-art summarization models \cite{wu2021recursively, zhang2021summ} on two datasets: movie screenplays \cite{chen2021summscreen}  and books \cite{kryscinski2021booksum}. Our work is the first to characterize specific errors made by these systems and gaps that exist with human-written coherent summaries. While recent efforts have studied errors in open-ended generation \cite{dou2021scarecrow}, these differ drastically from summarization errors and their taxonomies and findings are not transferable (see Appendix~\ref{appendix:open-ended-gen}).

We also evaluate the performance of automatic coherence models, comparing synthetic data generation techniques \cite{moon2019unified, shen2021evaluating} against \framename~annotations as training sources. Not only do models fine-tuned on \framename~outperform those trained on synthetic datasets, we find that they also report higher recall than individual human annotators at identifying fine-grained coherence error categories.


Our collected dataset and analysis provides a foundation for downstream applications such as better long summary evaluation, coherence-aware generation, and post-correction of summaries.

\section{Long Narrative Summarization}
\label{sec:dataset}
We study coherence errors in two domains, books and movie screenplays, although our taxonomy and annotation methodology are broadly applicable. 

\paragraph{Books} We evaluate the depth 1 book summaries generated by a GPT-3 based model \cite{wu2021recursively}. We evaluate both its 175B and 6B versions, denoted by \textsc{Book-175B} and \textsc{Book-6B} respectively. On average, these are \textasciitilde35 sentences long.
\paragraph{Movie Screenplays} We generate summaries for the movie scripts dataset \cite{papalampidi2020screenplay} using the \textsc{Bart}-based Summ\^{}N model \cite{zhang2021summ}.\footnote{Summ\^{}N is trained on TV episode screenplays. However, TV episodes are not self-contained narratives and often refer to events from previous episodes, making this an update summarization task which is harder to evaluate for coherence out of context. Therefore, we summarize movie scripts instead.} These are \textasciitilde40 sentences in length. We refer to them as \textsc{Movie-Bart}. 

The majority of prior research in evaluation of evaluation metrics \cite{kryscinski2019neural,bhandari2020re,fabbri2021summeval} has focused on news summarization datasets \cite{nallapati2016abstractive, narayan2018don}. However there exist substantial differences in the scale of news settings and narrative summaries: the former are considerably shorter at \textasciitilde3 sentences per summary. We first explore whether existing approaches to evaluation can work well despite this difference. 




\paragraph{Limitations of Current Human Evaluation}
\label{sec:limitations-human}
Summary-level Likert scale annotations are the most commonly used setup for collecting coherence in single-document news summarization research \cite{fabbri2021summeval}. Here, we run an analogous study for our longer narrative summaries. 

We ask 3 Mechanical Turk workers with prior experience in annotation for NLP tasks, specifically discourse analysis and text simplification, to rate the overall coherence of 100 generated summaries on a 5-point scale. Table \ref{tab:likert} reports the observed agreement, measured by Kripendorff's $\alpha$. Compared to newswire summaries collected under a similar setup \cite{fabbri2021summeval}, annotations for longer narratives have a much lower agreement. This shows the difficulty in obtaining a  consensus on coherence for a 500+ word summary through a single value on a 5-point scale.

\begin{table}[t]
    \centering
    \small
    \begin{tabular}{cc|c}
    \toprule
        News (Expert) & News (Crowd) & Books (Crowd) \\ \midrule
         0.41$^{**}$ & 0.48 & 0.19 \\
    \bottomrule
    \end{tabular} 
    \caption{Summary-level agreement, measured by Krippendorff's $\alpha$. $^{**}$Expert agreement after one round of annotations; this aligns with the crowd setting.} 
    \label{tab:likert}
\end{table}

In Appendix \ref{appendix:automatic-metrics-corruptions}, we further show that automatic metrics like \textsc{Rouge} and BERTScore \cite{zhang2019bertscore} that are primarily used for evaluating long document summarization fail to penalize coherence errors in summaries. Better tools for both automatic and human evaluation are needed. 

\section{\framename~Annotation Methodology} 
We design our methodology to: 1) simplify the summary-level annotation task into smaller sub-tasks, and 2) provide a structured
framework that allows annotators to specify the \textit{type} of coherence error, instead of evaluating coherence holistically.

\subsection{Task Workflow and Notation}
\label{sec:workflow}
We decompose the summary-level task into smaller segment-level tasks: at each step, annotators evaluate a subpart of the summary, which is usually 2-4 sentences long. Let ${S_0, S_1 ... S_N}$ denote these summary segments. While evaluating segment $S_{i}$, coherence judgments are made with respect to both the context ${S_0, S_1 ... S_{i-1}}$ and text within $S_{i}$. 

To annotate a single error in $S_i$, annotators select the error span $t_j \in S_i$ and the coherence error type $e_j$ (error taxonomy outlined in Section~\ref{sec:error-taxonomy}) to construct the error triple $a_j = (S_i, t_j, e_j)$. This process is repeated until all errors in segment $S_i$ have been added, after which they proceed to the next segment $S_{i+1}$ for annotation. At the end of the annotation, workers produce the full set of annotations $A = \{a_j \; \forall j\}$ across all the text segments. The outcome of this is shown in Figure~\ref{fig:error_types}.


For book summaries, i.e. \textsc{Book-175B} and \textsc{Book-6B}, our segments come from boundaries present in the generated summaries. These are an average of 2.7 sentences. For \textsc{Movie-Bart}, we segment summaries into chunks of 3 sentences.

\subsection{Error Taxonomy}
\label{sec:error-taxonomy}
\citet{reinhart1980conditions} states three conditions for coherence: connectedness (cohesion), consistency, and relevance. Our error taxonomy is guided by these conditions while covering the broad range of coherence errors produced by current models. 

We divide errors into two categories: a) \textbf{Coherence errors}: these measure whether the summary is well-structured and events in the summary make narrative sense, and b) \textbf{Language errors}: these measure other aspects of the quality of generated text, such as grammar. While these do not come under the ambit of coherence errors, we found it useful to provide these additional error types for crowd workers to anchor other ``badness'' in text to.  

\begin{figure*}[t]
    \centering
    \includegraphics[trim=15mm 145mm 35mm 20mm,scale=0.25, clip]{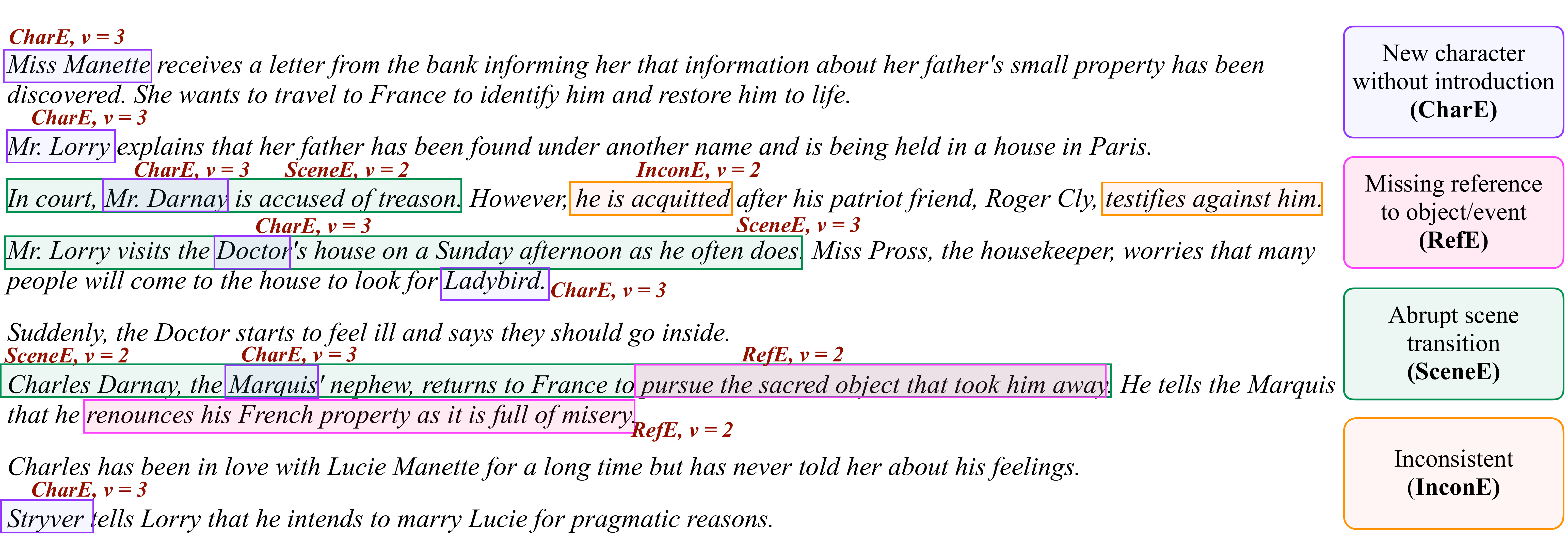}
    \caption{An example of expert annotations for a \textsc{Book-175B} summary (we only show coherence errors). The number of annotators who identified each span is denoted by $v$; for simplicity, we omit $v=1$ errors. We see that annotators often identify overlapping error spans; this fine-grained picture of coherence cannot be achieved by a summary-level score.}
    \label{fig:expert-annotations}
\end{figure*}

\subsubsection{Coherence Errors}
\paragraph{New character without introduction (\texttt{CharE})} These refer to scenarios where a new person is introduced in the narrative without providing any background about the person, or their relation with other characters in the story. This violates condition 1 of coherence, i.e. connectedness. Note that well-known people, e.g. Barack Obama, do not need an introduction.\footnote{We special-cased this class of error because it was so frequent in our data. Our narratives are about fictional people in real-world settings, so places, organizations, and other named entity types are less likely to require explicit introduction.}

\paragraph{Missing reference to an event or object (\texttt{RefE})} These refer to scenarios where an event or object is mentioned for the first time, but the phrasing strongly implies that it must have been introduced previously or that some context is missing to fully understand it. E.g., in Figure~\ref{fig:error_types}, the phrasing of \textit{her husband's suicide} gives the strong impression that the reader is already aware of this event. 

\paragraph{Abrupt scene transition (\texttt{SceneE})} These occur where there is a sudden shift in the narrative and are related to both connectedness and relevance. 
For these, we ask annotators to select whole sentences.

\paragraph{Inconsistency (\texttt{InconE})} These errors violate the second condition of coherence, i.e. contradicting other information in the \textit{Context} or within the \textit{Current Segment}. For these errors, we also ask annotators to choose the previous span it is inconsistent with.

\subsubsection{Language Errors} 
\paragraph{Repetition (\texttt{RepE})} These are used to detect repetition. Similar to \texttt{\textbf{InconE}}, annotators also select the antecedent span with the repeated content.

\paragraph{Ungrammatical or Nonsensical Text (\texttt{GramE})} These refer to text spans with grammar errors. Also included in this category are cases where there are obvious model degenerations.

\paragraph{Unclear coreference (\texttt{CorefE})} These refer to cases where it is unclear who or what a pronoun is referring to. While sometimes requiring extra clarity, we found that there errors rarely affected the overall narrative understanding  unless they co-occured with \texttt{\textbf{GramE}}. Therefore, we do not include them in the coherence error category.

The version of definitions and task instructions given to the annotators is in Appendix \ref{appendix:error-schema}.

\begin{table}[t]
\renewcommand{\tabcolsep}{1.8mm}
    \centering
    \small
    \begin{tabular}{c|c|c|ccc}
    \toprule
    Type & Dataset & \#summ & Span & Sent & Seg \\ \midrule
    \multirow{2}{*}{Expert} & \textsc{Book-175B}  &  5 & 323 & 173 & 111 \\
    &\textsc{Book-6B} & 5 & 401 & 174 & 66 \\ \midrule
    \multirow{3}{*}{Crowd} &\textsc{Book-175B} &  55 & 3.1k & 2.2k & 1.1k\\
    &\textsc{Book-6B} &  55 & 2.9k & 2.2k & 0.7k\\
    &\textsc{Movie-Bart} & 40 & 2.8k & 1.8k & 0.6k\\ \midrule
    \multicolumn{2}{c|}{Total}  & 160 & 9.6k & 6.6k & 2.6k\\
    \bottomrule
    \end{tabular}
    \caption{Statistics for expert and crowd annotations per level of granularity: span-, sentence- and segment-levels. Span-level annotations are multi-class, sentence- and segment-level have binary labels of coherence.}
    \label{tab:dataset-statistics-annotations}
\end{table}

\section{Data Collection} 
We collect annotations from two types of annotators: experts and crowdworkers. 

\subsection{Expert Annotations}
Expert annotations were collected from 3 authors who have previously published papers in text summarization and have experience engaging with model-generated text. Each annotator evaluated 10 book summaries, 5 each from \textsc{Book-175B} and \textsc{Book-6B}. This resulted in a dataset of \textasciitilde700 span-level error annotations. Furthermore, we project span-level annotations to obtain binary coherent (no coherence error) and incoherent labels (at least one coherence error) at the sentence- and segment-levels. Table \ref{tab:dataset-statistics-annotations} provides statistics at these levels.

We observed high inter-annotator agreement for expert annotators at both the sentence- and segment-levels (see Table~\ref{tab:agreement}). We used this dataset to train crowdworkers in the next stage.

\begin{figure*}[t]
    \centering
    \includegraphics[trim=15mm 205mm 25mm 35mm,scale=0.25,clip]{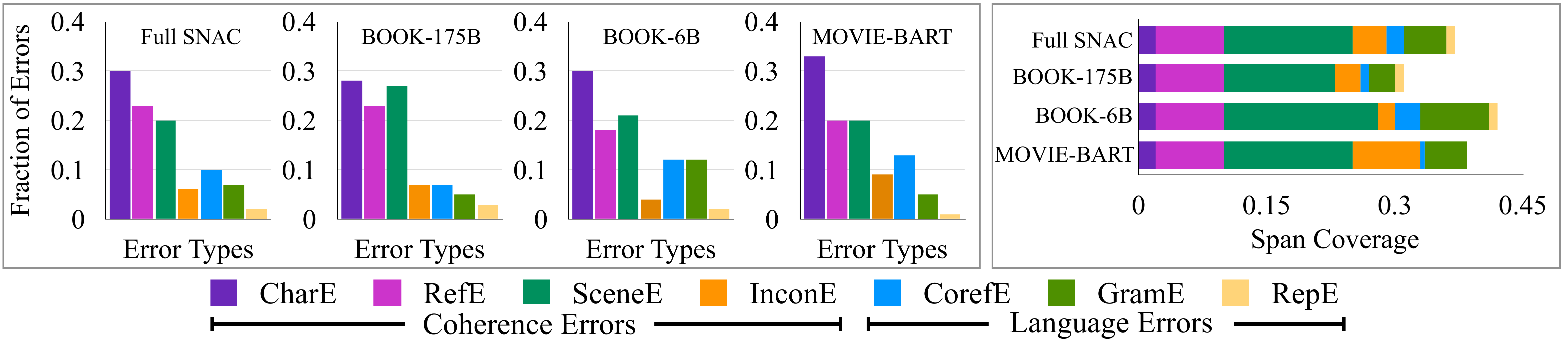}
    \caption{The left graph shows the fraction of times a specific error type is detected for each individual dataset: \texttt{\textbf{CharE}}, \texttt{\textbf{RefE}} and \texttt{\textbf{SceneE}} errors constitute the majority of coherence errors. The right graph shows the average fraction of tokens belonging to each error-type.}
    \label{fig:error_dist}
\end{figure*}

\subsection{Crowd Annotations}
\label{sec:crowd-workers}

We first launched a qualification task to recruit MTurk workers. The qualification was only made available to a subset of workers who had previously worked on other data annotation efforts for NLP tasks. It included detailed instructions explaining the task workflow, interface, and error schema. Each worker was asked to annotate 2 book summaries; these summaries were chosen from the set of expert annotations. Workers were paid \$12 for attempting this qualification. 

We evaluated each worker's annotations against experts and sent individual feedback. Among coherence errors, we observed that workers generally tended to disagree on \texttt{\textbf{RefE}}; each worker had a different calibration of which events or objects require more context to improve overall narrative understanding. Another common source of disagreement between workers and experts were \texttt{\textbf{SceneE}} errors. To help align their understanding with experts, we provided workers with a complete set of expert annotations for a whole summary for reference.

We recruited 11 workers after the qualification to annotate 150 generated summaries. Each summary was annotated by 3 different annotators. Workers were paid an average of \$12/hr.

\subsection{\framename~Dataset}
\label{sec:narcos-dataset}
Our resulting dataset consists of \textasciitilde9.6k span-level annotations for coherence judgments, across 160 summaries. 
Dataset statistics for the entire collected dataset, including both expert and crowd annotations, are shown in Table \ref{tab:dataset-statistics-annotations}. 

A summary-wide expert annotation is \framename{} is shown in Figure \ref{fig:expert-annotations}. Noticeably, \texttt{\textbf{CharE}} spans constitute the majority of errors; this observation is consistent throughout all datasets. Annotators tend to show higher recall and agreement over this category. \texttt{\textbf{SceneE}} and \texttt{\textbf{RefE}} are the next two major error categories. The annotations also illustrate the two reasons for \texttt{\textbf{SceneE}}: 1) there is a sudden change in setting and characters, e.g. \emph{Mr Lorry visits the...} and 2) the previous scene is abruptly cut off, e.g. \emph{In court, Mr. Darnay...}, where Ms.~Mannette's story is unfinished.  

We observed that worker annotations are high precision but low recall (\texttt{\textbf{CharE}} errors are an exception; workers have both high precision and recall for this category). This means that error spans identified by each worker tended to be actual errors, even when they were not detected by other annotators. Therefore, we combine annotations of all 3 annotators to construct the full \framename~dataset. 


\paragraph{Error Distributions}  Figure \ref{fig:error_dist} shows the fraction of unique errors of each error type annotated across all datasets. As seen in Figure \ref{fig:expert-annotations} annotations, the majority of the coherence errors are due to \texttt{\textbf{CharE}}, \texttt{\textbf{RefE}} or \texttt{\textbf{SceneE}}. The bottom graph of Figure~\ref{fig:error_dist} shows the number of error tokens annotated (instead of numbers of errors) for each error type. We see that annotators mark a larger fraction of tokens in the \textsc{Book-6B} dataset as erroneous compared to \textsc{Book-175B}. The main difference comes from the difference in \texttt{\textbf{SceneE}} (annotators are instructed to select entire sentences) and \texttt{\textbf{GramE}}. As expected, for smaller summarization models, i.e. \textsc{GPT-3 6B} and \textsc{Bart}, a larger fraction of errors and error tokens are associated with language errors compared to \textsc{GPT-3 175B}. In fact, we noticed that workers were more likely to skip coherence error annotations, e.g. \texttt{\textbf{RefE}}, when these co-occur with \texttt{\textbf{GramE}} for these models, particularly on \textsc{Book-6B}.

\paragraph{Human annotators focus on language errors while assessing coherence holistically.} To understand which aspects of a summary contribute to the summary-level coherence rating provided by crowd workers, we compute the correlation between the number of errors of each type with the overall coherence score (Likert rating on a scale of 1-5, described previously in Section~\ref{sec:limitations-human}).\footnote{We previously showed that annotators do not agree on overall summary ratings. However, this experiment differs in that each annotator's aggregated segment-level errors are correlated with \emph{their own} summary-level judgment; here, agreement between annotators is not relevant.} 

\begin{table}[t]
    \centering
    \small
    \begin{tabular}{r|c|r|c}
    \toprule
    Error Type & Coherence & \multicolumn{1}{c|}{Language} & Total \\ \midrule
    $r$ & -0.26$^*$ & \multicolumn{1}{c|}{\textbf{-0.34$^*$}} & -0.33$^*$ \\ \midrule 
    \multicolumn{2}{c|}{\cellcolor{Gray}Coherence Errors} & \multicolumn{2}{c}{\cellcolor{Gray}Language Errors}  \\ \midrule
    \texttt{\textbf{CharE}}  & -0.22$^*$ & \texttt{\textbf{RepE}} & -0.21\phantom{$^*$} \\
    \texttt{\textbf{RefE}}  & \textbf{-0.29$^*$} & \texttt{\textbf{CorefE}} & -0.24$^*$\\
    \texttt{\textbf{SceneE}} & -0.05\phantom{$^*$} & \texttt{\textbf{GramE}} & \textbf{-0.25$^*$}\\ 
    \texttt{\textbf{InconE}} & -0.09\phantom{$^*$} & \\
    \bottomrule
    \end{tabular}
    \caption{Pearson correlation between no. of errors and summary-level coherence score for error categories. Annotators tend to focus on grammar instead of coherence-specific errors while assigning overall summary-score. $*$: p-value $<0.05$, according to a two-tailed test.}
    \label{tab:corr-likert-error}
\end{table}

Table \ref{tab:corr-likert-error} outlines our results. First, it shows that the total number of errors is correlated with the overall coherence score, but annotators tend to weight language errors higher than coherence-specific errors. Surprisingly, we see negligible correlation with \texttt{\textbf{SceneE}} errors although these are a prominent distinguisher between generated and human-written summaries. Amongst other error types, both \texttt{\textbf{RefE}} errors and \texttt{\textbf{GramE}} errors show relatively higher correlation. Although not directly evaluating coherence, \citet{clark2021all} report similar observations where annotators tend to focus on grammar errors while judging text quality.

\paragraph{Narrative Summarization $\ne$ Open-Ended Generation} In story completion, models are not required to cover all salient information from a document and only condition on past generated text; generated open-ended summaries rarely diverge
off-topic. Examples of GPT-3 generated stories in
Figure \ref{fig:open-ended-generation} (Appendix~\ref{appendix:open-ended-gen}) show that these generate almost no \texttt{\textbf{CharE}}, \texttt{\textbf{RefE}} or \texttt{\textbf{SceneE}} errors that form the majority in \texttt{SNaC}, and instead mainly exhibit repetition. Therefore, research efforts that introduce fine-grained taxonomies for this task, e.g. \texttt{Scarecrow} \cite{dou2021scarecrow}, are directly applicable to
summarization which needs to be independently
studied.

\subsection{Inter-Annotator Agreement}
We first compute inter-annotator agreements at the \textbf{sentence- and segment-levels}. This allows for an apples-to-apples comparison with \citet{fabbri2021summeval} as the average length of news summaries is roughly equal to our segment length. We convert their 5-point Likert ratings into binary labels using the threshold that gives the best agreement score. We compare Krippendorff's $\alpha$ for \framename~and news in Table~\ref{tab:agreement}: \framename~reports high inter-annotator agreement at both the sentence- and segment-level. Notably, this segment level agreement is better than that of crowdworkers in the news domain. 

\begin{table}[t]
    \centering
    \small
    \begin{tabular}{r|cc|cc|c}
    \toprule
        & \multicolumn{4}{c|}{Our Annotations} & Newswire \\  \midrule
        & \multicolumn{2}{c|}{Expert} & \multicolumn{2}{c|}{Crowd} & Crowd \\
        & Sent & Seg & Sent & Seg & Seg \\ \midrule
        Coherence & .77 & .90 &.59 & .69 & .49 \\
        Language & .33 & .45 & .22 & .28 &  - \\ \bottomrule
    \end{tabular}
    \caption{Segment and sentence-level agreement, measured by Krippendorff's $\alpha$ for \framename. Our dataset reports higher inter-annotator agreement compared to newswire summaries adapted to a similar setting.}
    \label{tab:agreement}
\end{table}

\begin{table}[t]
    \centering
    \small
    \begin{tabular}{r|cc|cc}
    \toprule
        Error & \multicolumn{2}{c|}{Krippendorff's $\alpha$} & \multicolumn{2}{c}{Two-agree \%} \\
        & Expert & Crowd & Expert & Crowd \\ \midrule
        \texttt{\textbf{CharE}} & .91 & .69 & 86 & 67 \\
        \texttt{\textbf{SceneE}} & .57 & .30 & 62 & 35 \\
        \texttt{\textbf{RefE}} & .25 (.39) & .10 (22) & 27 (39) & 11 (23)\\
        \texttt{\textbf{InconE}} & .18 (.29) & .13 (21) & 20 (37) & 14 (23)\\
    \bottomrule
    \end{tabular}
    \caption{Token-level agreement for errors in the coherence sub-category. For \texttt{\textbf{RefE}} and \texttt{\textbf{InconE}}, we also report agreement (in parentheses) after normalizing span boundaries for overlapping errors.}
    \label{tab:agreement-token}
\end{table}

\paragraph{Span-level analysis} Next, we evaluate category-specific agreement between annotators at the span level. We report two metrics: 1) Krippendorff's $\alpha$ and 2) two-agree \%; borrowed from \citet{dou2021scarecrow}, this reports the percentage of tokens labeled as erroneous by at least one annotator that were also labelled by one or more additional annotators. For \texttt{\textbf{RefE}}  and \texttt{\textbf{InconE}}, we noticed that small differences in span boundaries caused a significant drop in agreement, therefore, for these we also report metrics after normalizing span boundaries of overlapping spans to their union.

Table \ref{tab:agreement-token} outlines the agreement: for both expert and crowdworkers, we see high agreement for \texttt{\textbf{CharE}} and fair agreement for \texttt{\textbf{SceneE}}. On the other hand, lower agreement is observed for \texttt{\textbf{RefE}}; this aligns with our observation that individual annotators may have low recall. Different annotators fundamentally have different notions of what extra information is critical for understanding the text.

Similar overall results at the token-level are reported by \citet{dou2021scarecrow} for their error taxonomy: their error categories \textit{Commonsense} and \textit{Encyclopedic} report the lowest metrics, the two-agree \% is as low as 20 and 12 respectively for 10 annotators. Note that we only have 3 annotators, so we expect our two-agree numbers to be much lower.\footnote{We omit comparison with Krippendorff's $\alpha$ reported in \citet{dou2021scarecrow} as they report \textit{observed} agreement without normalizing by \textit{expected} agreement. We re-compute their interannotator agreement on their dataset with normalization for a randomly selected subset of 3 annotators (comparable to our setting). This gives an average of 0.14 Krippendorff's $\alpha$ across all categories, with the bottom 5 categories reporting an average of 0.05 $\alpha$.} 

\begin{figure}[t]
    \centering
    \includegraphics[trim=25mm 268mm 35mm 24mm,scale=0.22,clip]{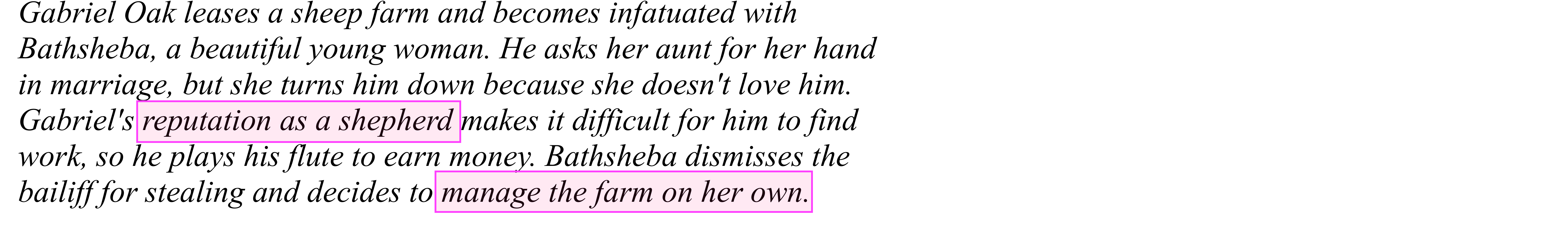}
    \caption{\texttt{\textbf{RefE}} errors identified by only one annotator.}
    \label{fig:event-low-agreement}
\end{figure}

Figure \ref{fig:event-low-agreement} shows an example of a summary with low crowd agreement over the \texttt{\textbf{RefE}} errors (we omit all other identified errors in this figure). For the first highlight, it is reasonable to seek more clarity on why \textit{Gabriel's reputation as a shepherd makes it difficult for him to find work} as it presupposes negative connotations associated with his profession that the reader is not privy to. The second highlight asserts that \textit{Bathsheba} owns or works at ``the farm'' as a known fact, which is information that has  not been mentioned previously. Although only annotated by one annotator, these qualify as \texttt{\textbf{RefE}} errors according to our definition.

\section{Detecting Coherence Errors} 
\label{sec:benchmarking}



\paragraph{Setup} 

We aim to see whether models can automatically detect our coherence errors. We formulate all models as sequence classifiers: given a context $c$ and a sentence $s$, the goal is to classify whether $s$ contains coherence errors. Similar to Section \ref{sec:narcos-dataset}, we project span-level errors to a sentence-level gold coherence label $y^* \in \{0,1\}$. Let $E=\{(e_j^*, t_j^*)\}$ denote the set of error types and corresponding spans in $s$.

We split \framename{} into train (4.2k), dev (230) and test (1.8k) examples and evaluate on the test set.

\paragraph{Metrics}
First, we consider a \textbf{sentence-level binary classification} version of this task: can models correctly predict if a sentence contains coherence errors? In this case, our models take the form $P(y \mid c, s)$ where $y \in \{0,1\}$. We report precision, recall and F1. Note that the sentence-level $y^\textrm{pred}$ judgment can be due to any of the error types. 

We next evaluate \textbf{fine-grained prediction}: can models identify the specific coherence error type and pinpoint the error span? In this case, our models predict $P(\mathbf{y} \mid c, s)$, where $\mathbf{y}$ is a bundle consisting of $y$ and a set of error tuples $\{(e_j^\textrm{pred}, t_j^\textrm{pred})\}$ if $y = 0$. We report the precision, recall and F1 performance at correctly identifying the error type, i.e. $e_j^\textrm{pred} = e_j^* \; \forall e_j$. We also report \texttt{ov.} computed as the fraction of times the predicted error span overlaps with the correct error span.

\subsection{Models for Comparison}
We compare performances of three types of models: (1) unsupervised (UNSUP). (2) Models trained on synthetic data targeting coherent errors (SYN). We follow prior work \cite{joty2018coherence,shen2021evaluating} and generate synthetic training data by introducing artificial coherence errors in reference text, specifically on the BookSum dataset \cite{kryscinski2021booksum}. We ensure zero overlap between this synthetic train set and the evaluation test set. (3) Models fine-tuned on the \framename~data (FT). 

\paragraph{(UNSUP) LM Perplexity} We use GPT-2 \cite{radford2019language} to obtain the probability of the sentence $s$, given the context $c$, i.e. $P(s \mid c)$. The dev set is used to select a threshold $\tau_{LM}$ and obtain binary labels from these probabilities: predict an error if $P(s \mid c) < \tau_{LM}$. 

\paragraph{(UNSUP) Entity Grid} We construct entity grids \cite{barzilay2005modeling, barzilay2008modeling} for both predicted and gold summaries in order to compare their discourse structures. Using gold summaries in the BookSum dataset, we estimate the probabilities of syntactic role transitions of entities between sentences, e.g. $p(S \rightarrow O)$, $p(S \rightarrow X)$, $p(O \rightarrow S)$, etc. Then, we score the coherence of a predicted summary $s$ as the log probability of the transition from $c^{-1}$, i.e. the last sentence of context $c$, to sentence $s$: $w(c, s) = \sum_{e \in E} \log p(r(s, e) \mid r(c^{-1}, e))$. Here, $E$ is the full set of entities in $s$ and $c^{-1}$ and $r(x, e)$ denotes the role of entity $e$ in sentence $x$.

The \framename~dev set is used to select a threshold $\tau_{EG}$ and obtain binary labels from these scores: predict a coherence error if $w(c, s) < \tau_{EG}$.

\begin{figure}[t]
    \centering
    \includegraphics[trim=100mm 55mm 40mm 46mm,scale=0.23,clip]{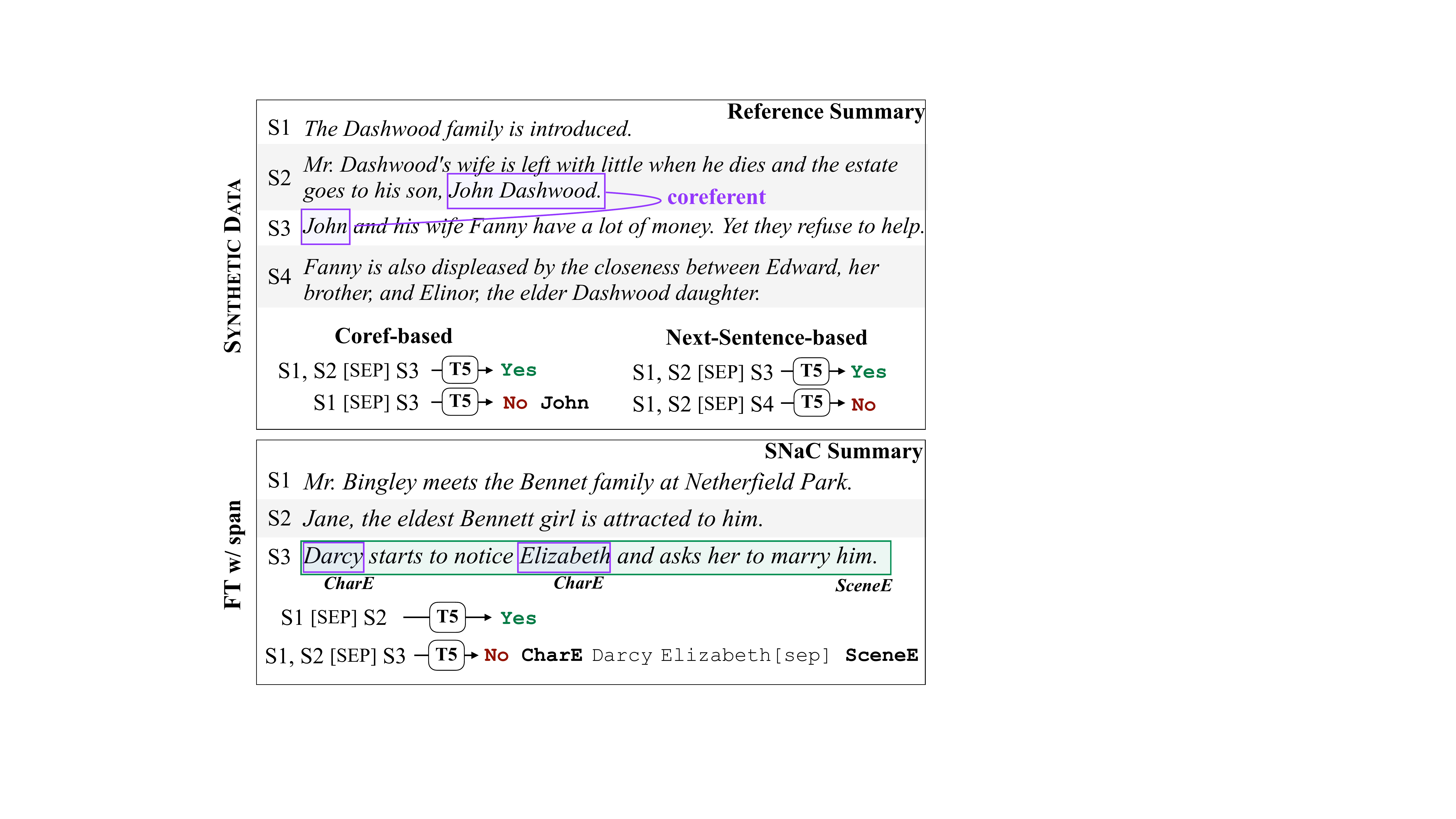}
     \caption{Training data generation, and T5 inputs and outputs for the SYN and FT w/ span models. The FT w/o span model only generates yes/no and not the specific category or span.} 
    \label{fig:training-data-gen}
\end{figure} 
 
\paragraph{(SYN) Coref-based} This technique is designed to specifically target \texttt{\textbf{CharE}} and \texttt{\textbf{RefE}} errors. 
We run a coreference model \cite{lee2018higher} to extract coreferent chains in gold summaries. Let $s_i$, $s_{j>i}$ be sentences with the first and second mention of an entity. We derive non-coherent examples by setting $s=s_j$ and removing sentence $s_i$ from the context, i.e. $c= s_1 s_2...s_{i-1} s_{i+1}... s_{j-1}$ (shown in Figure \ref{fig:training-data-gen}). Conversely, for positive coherent training data, we retain the original context from the gold summaries, i.e. $c= s_1 s_2 ...s_i ... s_{j - 1}$. We fine-tune T5-Large \cite{raffel2020exploring} for binary classification $P(y \mid c, s)$ on these $(y, c, s)$ triples; training data sizes and intrinsic performance are reported in Appendix \ref{appendix:benchmarking-models}. 

\paragraph{(SYN) Next-Sentence} This method is designed to target \texttt{\textbf{SceneE}} errors and closely resembles the sentence insertion method from prior work \cite{shen2021evaluating}. 
Given context $c= s_1 s_2 ... s_i$, we obtain negative coherence examples by replacing the next sentence with another randomly sampled sentence from the remainder of the same summary, i.e. $s=s_j$, where $j > i + 1$. Positive examples are created by retaining the original summary completion, i.e. $s=s_{i+1}$. Figure \ref{fig:training-data-gen} illustrates this. We fine-tune T5-Large to model $P(y \mid c, s)$. 


\paragraph{(FT) Models trained on \framename~data} We consider two versions: 1) w/o span: trained to generate yes/no reflecting the coherence of sentence $s$, and 2) w/ span: trained to additionally predict the error category (e.g. \texttt{\textbf{CharE}}) and the corresponding error spans. Note that $s$ can have errors belonging to multiple error categories, the model is trained to generate these in sequence. Figure \ref{fig:training-data-gen} illustrates this. For \texttt{\textbf{SceneE}}, we omit span prediction as these are designed to incorporate the whole sentence. Similar to SYN, we fine-tune T5-Large on these datasets. 

\begin{figure}
    \centering
    \includegraphics[trim=0mm 0mm 10mm 15mm,scale=0.35, clip]{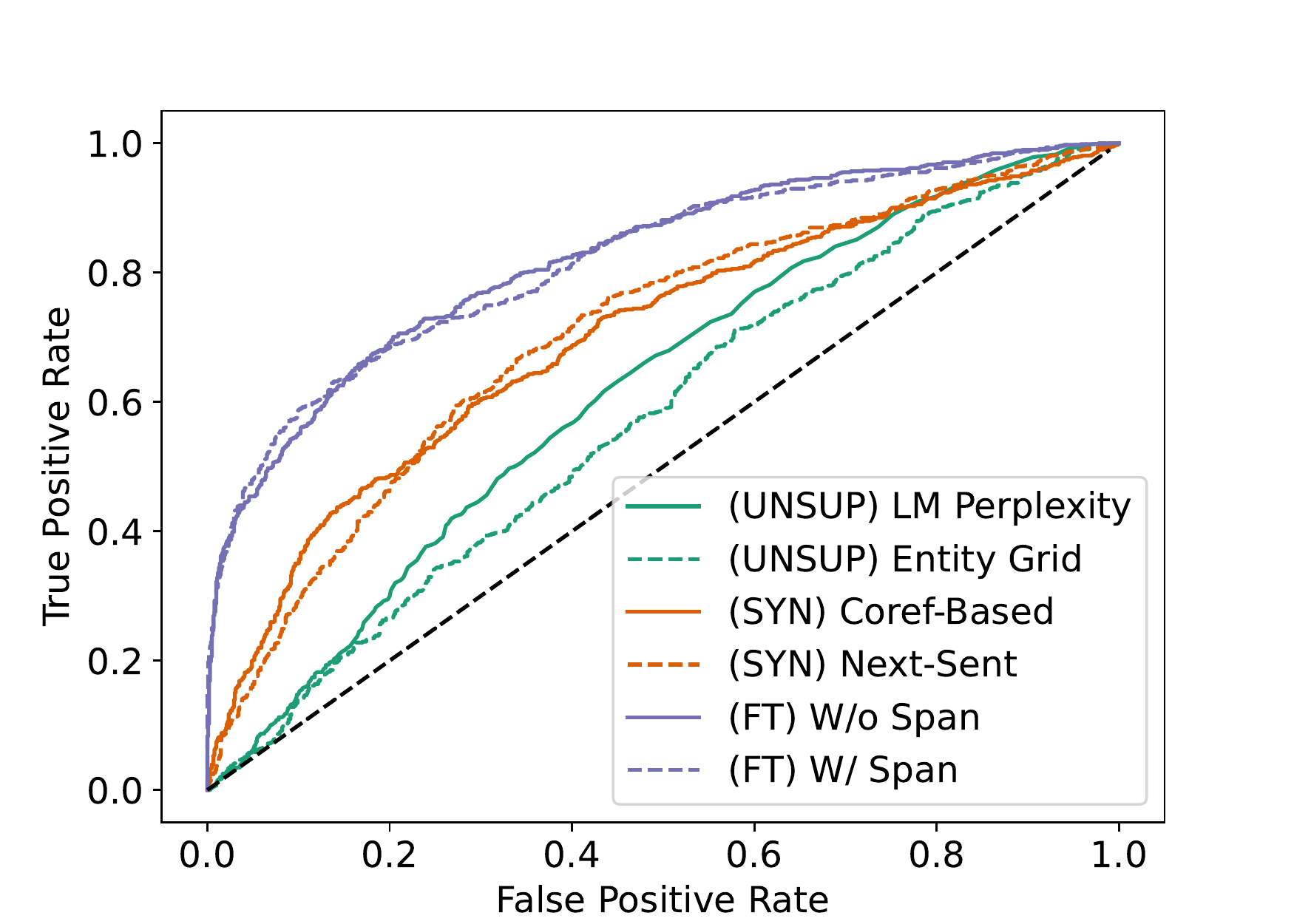}
    \caption{Performance of the different models on the \framename~test set. Models trained on \framename~ outperform those trained on synthetically generated datasets.}
    \label{fig:roc}
\end{figure}

\subsection{Results} 

\paragraph{Sentence-level binary classification}
Figure~\ref{fig:roc} shows an ROC curve of different models; the dotted black line indicates random chance. It shows that the entity-grid approach performs poorly compared to all neural approaches. Next, all trained models outperform the LM perplexity model; language models aggregating token-level probabilities cannot detect coherence errors. Finally, models trained on \framename~outperform synthetic datasets which are the primary source of training data in prior coherence work. This show that human annotations are needed to train strong coherence classifiers. 

\paragraph{Fine-grained prediction} Only our FT w/ span model is trained to predict both the error category and the corresponding spans. Therefore, we compare its performance against human annotators. For an apples-to-apples comparison, we re-construct our test set by aggregating annotations of two randomly chosen annotators. This unfairly penalizes FT w/ span by introducing a mismatch between its train and test conditions, especially precision. Therefore, we also report precision scores on the original test set in brackets. Full set of results on the original test set are in Appendix \ref{appendix:benchmarking-models}.

\begin{table}[t]
\renewcommand{\tabcolsep}{1.6mm}
    \centering
    \small
    \begin{tabular}{l|cccc|cccc}
    \toprule
        \multirow{2}{*}{\textbf{Error}} & \multicolumn{4}{c|}{\textbf{FT w/ span}} & \multicolumn{4}{c}{\textbf{Human}} \\
         & P & R & F1 & ov. & P & R & F1 & ov. \\ \midrule
        \texttt{\textbf{CharE}} & .79 (.86) & .81 & .80 & .98 & .88 & .71 & .79 & .98\\
        \texttt{\textbf{SceneE}} & .35 (.58) & .49 & .40 & 1.0 & .58 & .36 & .44 & 1.0\\
        \texttt{\textbf{RefE}} & .19 (.44) & .22 & .21 & .88 & .31 & .17 & .22 & .92\\
        \texttt{\textbf{InconE}} & .25 (.25) & .02 & .04 & 0.0 & .29 & .16 & .20 & .97\\
    \bottomrule
    \end{tabular} 
    \caption{Comparison between FT w/ span model and humans. Humans have higher precision while trained models report better recall across the top 3 error types.}
    \label{tab:localization}
\end{table}

Table \ref{tab:localization} outlines the results. As observed during qualitative evaluation, the held-out human annotations are high precision and low recall. On the other hand, FT w/ span is trained on the aggregated annotations from three annotators and reports higher recall than humans. Consequently, its F1 scores are comparable to human performance except for \texttt{\textbf{InconE}}. We attribute this to the limited number of training examples of this category.  

Similar to previous analysis, we observe that models and humans report the best performance at detecting \texttt{\textbf{CharE}}. Interestingly, the trained model can identify both \texttt{\textbf{SceneE}} and \texttt{\textbf{RefE}} with higher recall compared to human annotators. For these top three error types, trained models are successful at localizing error to specific spans, reporting high overlap scores. 

\section{Discussion}
Our analysis of current narrative summarization models \textbf{reveals that these do not generate coherent narratives}; in fact,  each generated summary contains \textasciitilde30 coherence errors of varying degrees of severity. Moreover, both automatic and human approaches for coherence evaluation fail to reliably measure coherence. \framename~addresses this gap.

However, we stop short of providing a prepackaged metric: which errors are more severe is application-dependent and subjective, and overall error counts cannot be compared. 
We encourage future work to focus on fine-grained error annotations, like those we present here, instead of sentence- or document-level annotations that do not provide actionable insights. We also recommend fine-grained error modeling for future coherence systems as well. While previous modeling has targeted document- or sentence-level coherence, our models trained on \framename~data can detect span-level coherence errors, particularly \texttt{\textbf{CharE}} errors with high accuracy. This automatic error localization opens up future avenues of post-hoc error correction systems built on top of coherence models. 

\section{Related Work}
\paragraph{Coherence frameworks} Inspired by Centering Theory \cite{grosz1995centering}, \citet{barzilay2005modeling, barzilay2008modeling} proposed entity-grid models to measure coherence through transitions of entity roles. This was further extended to incorporate non-head entities \cite{elsner2011extending}, discourse roles \cite{lin2011automatically}, and other improvements \cite{feng2012extending, feng2014impact}, including neural variations \cite{guinaudeau2013graph, nguyen2017neural, joty2018coherence} to better model text coherence. However, these models have been evaluated primarily on document-level essay scoring tasks \cite{mesgar2018neural} or artificial sentence-ordering tasks \cite{shen2021evaluating}, and not on model-generated coherence errors. 

\paragraph{Summarization Evaluation} Automatic metrics such as \textsc{BLEU} \cite{papineni2002bleu}, \textsc{METEOR} \cite{banerjee2005meteor}, \textsc{Rouge} \cite{lin2004rouge}, BERTScore \cite{zhang2019bertscore}, and others have been used to evaluate summarization, but \citet{fabbri2021summeval} showed that these correlate poorly with summary quality. Human evaluation is widely considered the gold standard for generation tasks, however, recent work \cite{karpinska2021perils, clark2021all} demonstrated that humans are not reliable for evaluating strong models like GPT-3. 

\section{Conclusion} 
We introduce \framename, a narrative coherence evaluation framework for long summaries. We develop an error taxonomy grounded in coherence errors made by current models and annotate data to provide the first characterization of such errors in narrative summaries. We also make our annotation tool publicly available to support future research efforts. 

\section{Limitations}

Although we view this work as an important step towards better understanding and evaluation of coherence in summaries, we acknowledge there is much more to do here. In this work, we only collect annotations and analyze coherence errors in summaries of English language books and movie screenplays. Our proposed taxonomy may not cover errors made by text summarization models for other languages and our trained models and analysis are English-specific. 

Moreover, some of these books summarized were written decades ago and may reflect the societal biases of those times, which could conceivably bias our trained error detection models. In this work, we use the text from the model generated summaries as is and do not perform any filtering.

Finally, our work studies generated summaries for long narrative text. While we believe that our taxonomy is generalizable to other types of narrative text, we do not investigate whether it covers other domains involving summarization of long documents, such as government report summarization \cite{huang2021efficient} or meeting summarization \cite{zhong2021qmsum}. 

\section*{Acknowledgments} Thanks to Eunsol Choi and Michael Strube for providing feedback on this work, as well as our Mechanical Turk annotators for conducting the annotation. This project was partially supported by Good Systems,\footnote{https://goodsystems.utexas.edu/} a UT Austin Grand Challenge to develop responsible AI technologies, a grant from the UT Austin Office of the Vice President for Research through the ``Creating Connections for National Security Research Grants'' program, a grant from Open Philanthropy, NSF grants IIS-2107524, IIS-2145479, and gifts from Salesforce, Amazon, and Adobe.

\bibliography{anthology,custom}
\bibliographystyle{acl_natbib}

\appendix

\section{Narrative Summarization $\ne$ Open-Ended Generation}
\label{appendix:open-ended-gen}

\begin{figure*}
    \centering
    \includegraphics[trim=105mm 36mm 135mm 30mm, scale=0.52, clip, angle=270]{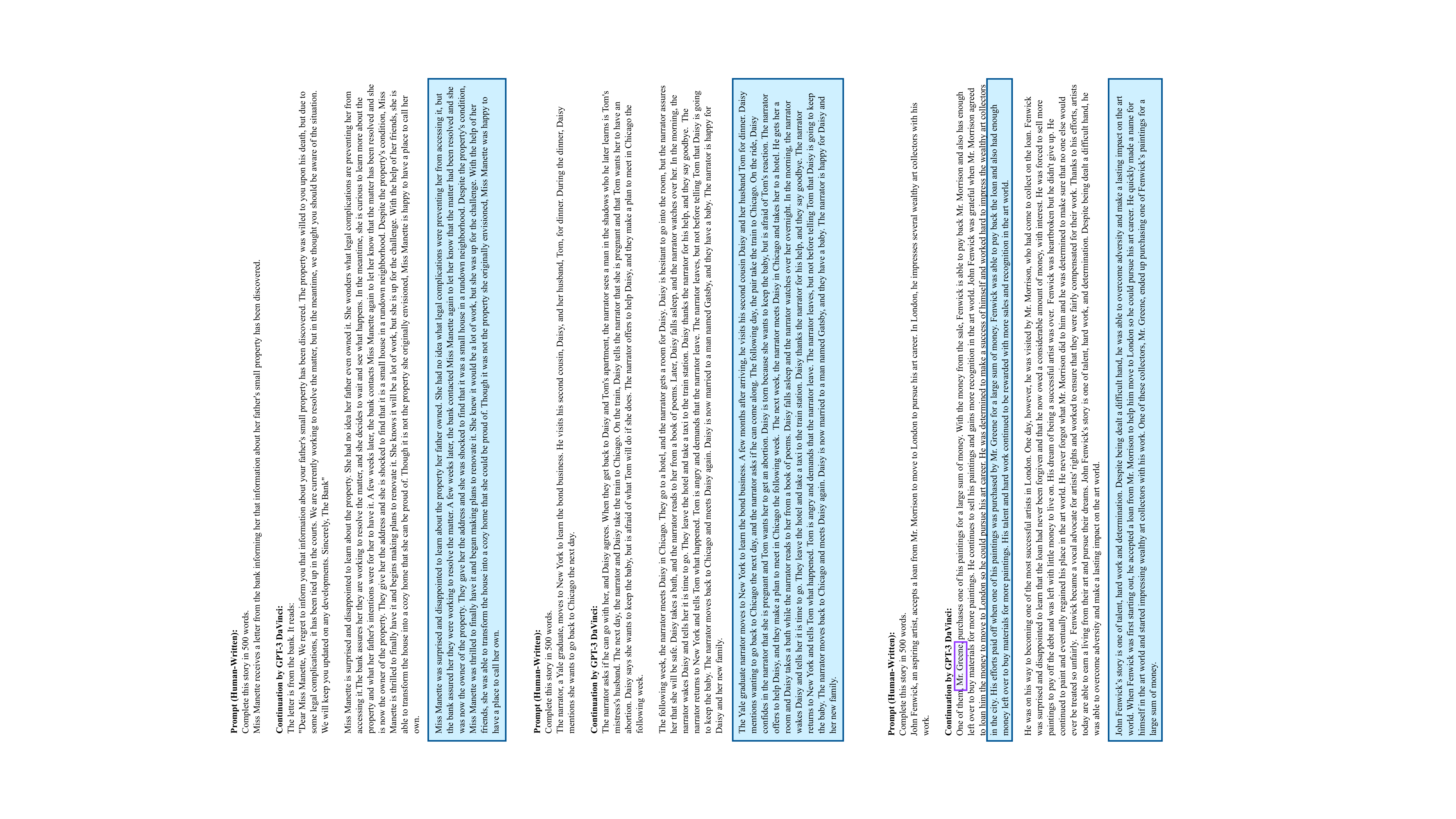}
    \caption{Examples of open-ended story completion by the GPT-3 text-davinci-002 model. The coherence errors observed under this setting (chiefly repetition errors, in blue) have little or no overlap with those from the summarization setting. Therefore, error taxonomies like \textsc{Scarecrow} that are devised for open-ended generated are not applicable to the summarization task.}
    \label{fig:open-ended-generation}
\end{figure*}

In Section~\ref{sec:narcos-dataset}, we noted that narrative summarization exhibits substantially different errors than open-ended text generation tasks like story generation or story completion, hence the need for our new taxonomy. We show examples of generated stories using the GPT-3 DaVinci in Figure~\ref{fig:open-ended-generation}. We prompt the GPT-3 text-davinci-002 model with the first few sentences of three generated summaries and ask for a 500-word completion. The coherence errors contained in these model outputs are very different from those in our narrative summarization setting. In particular, the stories hardly introduce any new characters (only \emph{Mr. Greene} is introduced in the third example), and when they do, these are properly contextualized with the narrative. Furthermore, these models rarely generate \texttt{\textbf{RefE}} and generate no \texttt{\textbf{SceneE}} type of errors. In fact, repetition errors, shown in blue, dominate these narratives. Therefore, error taxonomies devised for these tasks, e.g. \textsc{Scarecrow} \cite{dou2021scarecrow}, are not useful for summarization settings  which needs to be independently studied.

\section{Limitations of Automatic Metrics}
\label{appendix:automatic-metrics-corruptions}
Long document summarization research \cite{chen2021summscreen, huang2021efficient, kryscinski2021booksum, mao2021dyle, pang2022long} has primarily relied on \textsc{Rouge} scores to evaluate summaries. But do these capture narrative coherence? 

We test this for long narrative summaries, using the \textsc{Book-175B} dataset as a case study. Specifically, we test whether \textsc{Rouge} or BERTScore \cite{zhang2019bertscore} can differentiate between actual generated summaries and their corrupted versions with artificially injected coherence errors. We introduce 3 types of coherence errors to generated summaries: 
\begin{enumerate}[leftmargin=*]
    \item Random \textbf{shuffling} using a random permutation of all sentences in a \textsc{Book-175B} summary. This does not change the overall length of the generated summary.
    \item \textbf{Repetition} of a randomly selected subset of sentences. We randomly sample 50\% of the sentences to repeat, all other sentences only occur once.
    \item Retaining only \textbf{named entities} in the summary and \textbf{top generated bigrams}. We first extract the top 200 bigrams from the generated summaries in \textsc{Book-175B}, which include frequent bigrams like \textit{of the}, \textit{that he}, \textit{then he}, \textit{in the}, etc. For each test set summary, we construct a corrupted summary by concatenating all named entities in the summary (appending each named entity as many times as it occurs in the original summary) and the top bigrams extracted from the testset-wide summaries.
\end{enumerate}
For an upper bound, we also report metrics for a different human-written summary for the same input book sampled from the BookSum dataset.

\paragraph{Automatic metrics fail to penalize coherence errors.} Table \ref{tab:rouge-corruption} shows that both shuffling and repetition do not hurt \textsc{Rouge} or BERTScore, despite introducing critical coherence errors in generated summaries. The \textit{+NE \& bigram} setting does lead to a significant drop in BERTScore as these summaries are no longer fully-formed sentences. However, even this trivial baseline reports \textsc{Rouge} scores on par with the original \textsc{Book-175B} summaries, showing that \textsc{Rouge} is easy to ``game'' for this task. Finally, we see that human-written summaries, i.e., gold coherent summaries, only report 2 points of improvement in R2 and BERTScore over artificially incoherent baselines. This clearly shows that these metrics are inadequate to measure coherence, or even overall quality, for long summaries.

\begin{table}[t]
    \centering
    \small
    \begin{tabular}{r|cccc}
    \toprule 
    Summary & R1 & R2 & RL & BERTScore\\ \midrule
    OpenAI 175B & 41.9 & 11.0 & 17.1 & .51 \\
        + Shuffled & 41.9 & 11.0 & 15.6 & .51\\
        + Repetition & 44.7 & 10.6 & 17.2 & .49\\
        + NE \& bigram  & 42.8 & 10.1 & 16.3 & .26 \\ \midrule
    Human-written & 45.8 & 12.5 & 17.9 & .53\\ 
    \bottomrule
    \end{tabular}
    \caption{\textsc{Rouge} and BERTScore for \textsc{Book-175B} and several artificially corrupted versions. Results show that automatic metrics fail to penalize coherence errors.}
    \label{tab:rouge-corruption}
\end{table}

\section{Detecting Coherence Errors: Details and Additional Results}
\label{appendix:benchmarking-models}

\subsection{Models trained on synthetic data (SYN)}

Table \ref{tab:synthtic-intrinsic} shows the training data and development data size, as well as the intrinsic performance of these synthetic dataset-based coherence models on this development set. We construct both our datasets with an equal number of positive and negative coherence examples. The results show that T5 learns to model the synthetic task with reasonable accuracy. We do not expect the models to perform perfectly, as the synthetic data may have false positives (examples constructed to exhibit errors that are actually coherent).

\begin{table}[t]
    \centering
    \small
    \begin{tabular}{c|cccc}
    \toprule
        Method & \#train & \#dev & F1 & Acc. \\ \midrule
        Coref-based & 6.0k & 920 & .78 & .77 \\
        Next-Sent & 3.8k & 880 & .71 & .74 \\
    \bottomrule
    \end{tabular}
    \caption{Dataset sizes and intrinsic performance of T5-Large models trained on synthetic datasets.}
    \label{tab:synthtic-intrinsic}
\end{table}

\subsection{Implementation Details}

Table \ref{table:hyperparameters} shows the hyperparameters used for fine-tuning the T5-Large models on both synthetic training datasets and \framename.
\begin{table}[h]
    \small
    \begin{tabular}{l|l}\toprule
        Computing Infrastructure & 32GB NVIDIA V100 GPU \\
        Max Input Seq Length & 1024 \\
        Max Output Seq Length & 80 (for FT w/ span) \\
        Optimizer & Adam\\
        Optimizer Params & $\beta=(0.9, 0.999), \epsilon=10^{-8}$ \\
        Learning Rate Decay & Linear \\
        Learning rate & 1e-4\\
        Batch size & 8 \\
        Epochs & 5 \\ \bottomrule
    \end{tabular}
    \caption{Hyperparameters used for fine-tuning T5-Large on synthetic and \framename~ train sets.}
    \label{table:hyperparameters}
\end{table}

\subsection{Additional Results}
\label{sec:app_additional_results}

\paragraph{Sentence-level binary classification}
In Section \ref{sec:benchmarking}, we reported sentence-level binary classification results for all models. However, the sentence-level $y^\textrm{pred}$ judgment in that setting can be due to any of the 4 error types or their combination and binary classification metrics do not tell us which of these error types are easier to detect.

To answer this, we compute the error-wise recall under the binary setting. We assume $e^\textrm{pred}_{j} = 0$ if $y^\textrm{pred} = 0$ for all error types $e_j$; that is, a prediction of a binary error counts as detecting an error of any type in that sentence. This overestimates the recall performance and can be viewed as an upper bound; a model that can only detect \texttt{\textbf{CharE}} may report non-zero recall for other errors if these co-occur with \texttt{\textbf{CharE}}. 

\begin{table}[t]
\renewcommand{\tabcolsep}{1.8mm}
    \centering
    \small
    \begin{tabular}{r|ccccc}
    \toprule
        Model & \texttt{\textbf{CharE}} & \texttt{\textbf{SceneE}} & \texttt{\textbf{RefE}} & \texttt{\textbf{InconE}} & All\\ \midrule
        Coref-based & .61 & .47 & .48 & .15 & .43 \\
        Next-Sent & .31 & .35 & .32 & .09 & .27 \\ \midrule
        FT w/o span & .89 & .84 & .64 & .51 & .73\\ 
        FT w/ span & .90 & .82 & .58 & .47 & .70\\
    \bottomrule
    \end{tabular}
    \caption{Sentence-level recall of different errors types. Models (except FT w/ span) do not predict the error category; here, we treat these methods as binary classifiers and compute recalls as described in Appendix~\ref{sec:app_additional_results}.}
    \label{tab:category-wise-recall}
\end{table}

For fair comparison between different models, we report category-wise recall for all models at the same precision level $P=0.7$. Table \ref{tab:category-wise-recall} outlines our results. Both synthetic models report higher recall for the error category they were designed for. E.g., the coref-based method can detect \texttt{\textbf{CharE}} errors better than other error types. However, our FT models significantly outperform both synthetic approaches across all error types at thresholds with high precision performance. In particular, we observe high recall scores for \texttt{\textbf{CharE}} and \texttt{\textbf{SceneE}}.

\paragraph{Fine-grained prediction} In Table \ref{tab:localization}, we compared human and model (FT w/ spans) performance  on a modified test set created by combining annotations from 2 crowdworkers. This unfairly penalized the trained models, which may have slightly higher recall due to being trained on annotations from 3 crowdworkers. In Table \ref{tab:localization-full-narcos}, we report results on the original test set that combines annotations from all 3 annotators.

\begin{table}[h]
    \centering
    \small
    \begin{tabular}{r|cccc}
    \toprule
        Error & P & R & F1 & ov. \\ \midrule
        \texttt{\textbf{CharE}} & .86 & .74  & .80  & .99  \\
        \texttt{\textbf{SceneE}} & .58  & .49 & .53  & 1.0  \\
        \texttt{\textbf{RefE}} & .45  & .25 & .32  & .87 \\
        \texttt{\textbf{InconE}} & .25  & .01 & .02  & 0.0 \\
    \bottomrule
    \end{tabular}
    \caption{Performance of the T5-Large model fine-tuned on the \framename~dataset at predicting the correct error type in each summary sentence. We also report the percentage of times the predicted span overlaps with the error span in the gold data.}
    \label{tab:localization-full-narcos}
\end{table}

\section{\framename~Error Types and Task Interface}
\label{appendix:error-schema}

The definitions of error types and illustrative examples provided to the crowdworkers during training are outlined here.

\subsection{\texttt{\textbf{CharE}}}
\begin{figure}[h]
    \centering
    \includegraphics[trim=40mm 170mm 240mm 30mm, scale=0.18, clip]{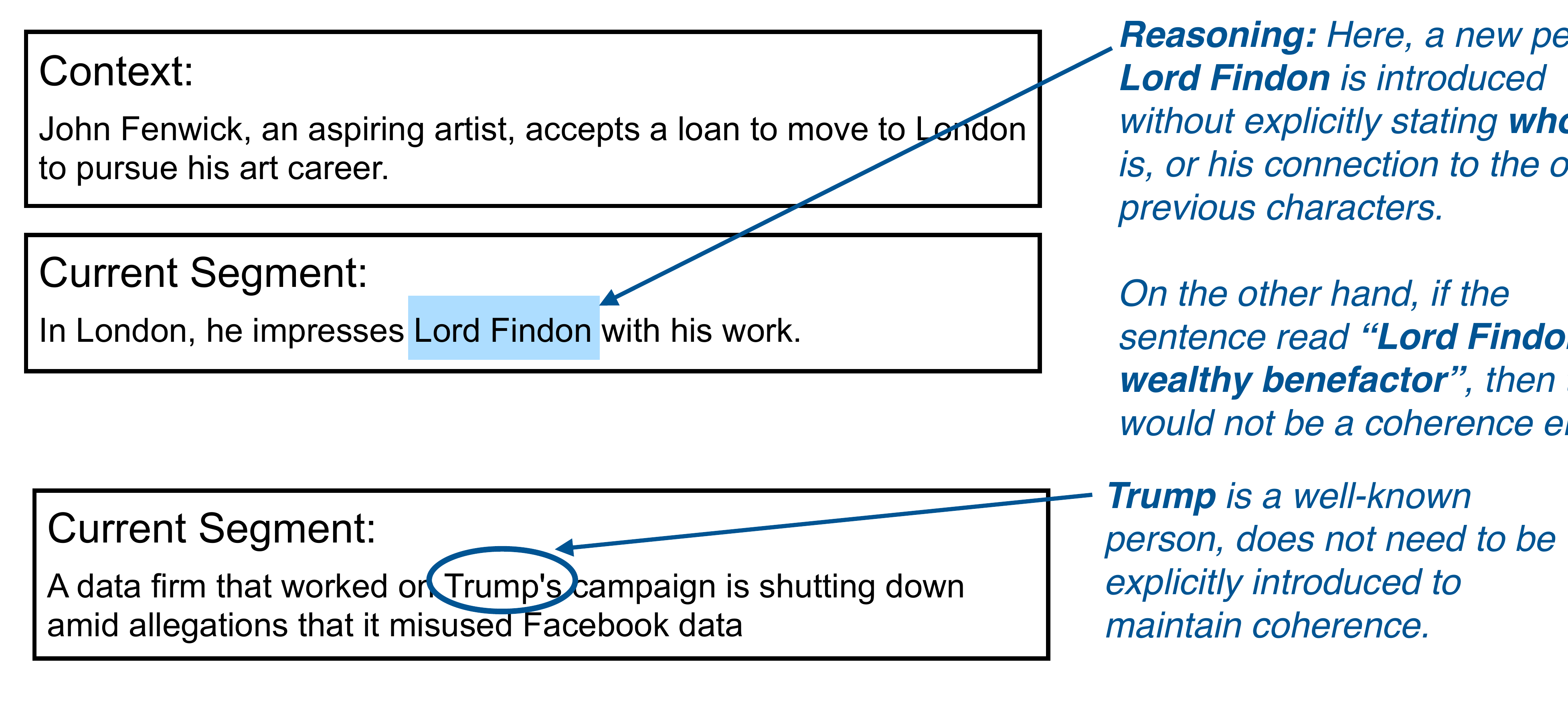}
    \caption{Illustration of \texttt{\textbf{CharE}} errors provided to crowdworkers during training.}
    \label{fig:charE}
\end{figure}

We call these \textbf{New Person not Introduced} in the task interface. We provide the illustrative example show in Figure \ref{fig:charE} along with the following definition:

``\textit{These refer to coherence errors where a new person is introduced into the narrative WITHOUT providing any background about the person, or their relation with other characters in the story. Note, however, that famous or well-known people do not need to be explicitly introduced.}''

\subsection{\texttt{\textbf{RefE}}}
\begin{figure}[h]
    \centering
    \includegraphics[trim=40mm 140mm 240mm 10mm, scale=0.20, clip]{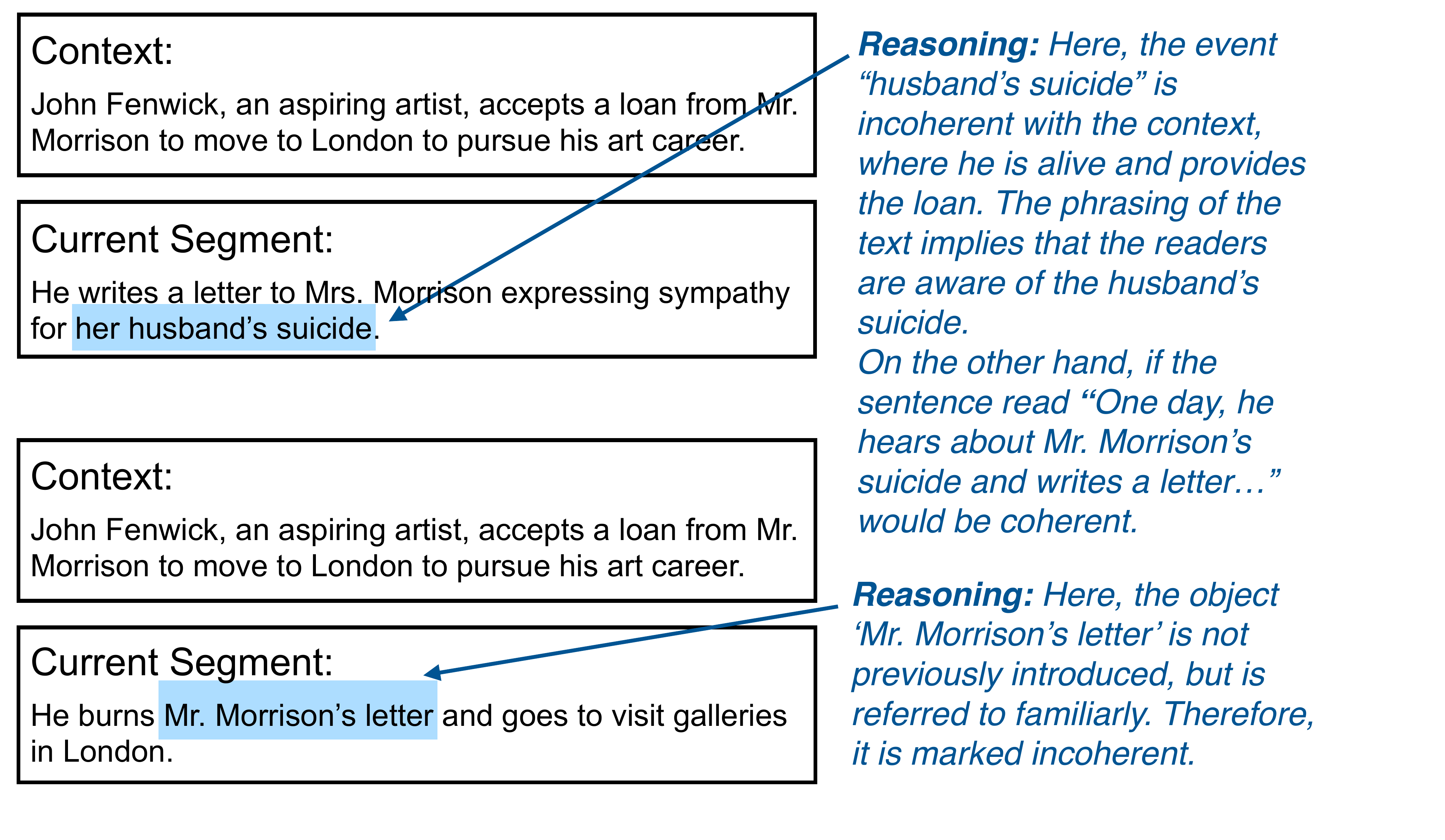}
    \caption{Illustration of \texttt{\textbf{RefE}} errors provided to crowdworkers during training.}
    \label{fig:refE}
\end{figure}

We call these \textbf{Missing Information about an Event/Object} in the task interface. We provide the illustrative example show in Figure \ref{fig:refE} along with the following definition:

``\textit{These refer to coherence errors where an event or object is mentioned for the first time, but the phrasing strongly implies some context is missing to understand this event/object and that it must have been introduced previously.}''

\subsection{\texttt{\textbf{SceneE}}}
\begin{figure}[h]
    \centering
    \includegraphics[trim=20mm 180mm 240mm 35mm, scale=0.25, clip]{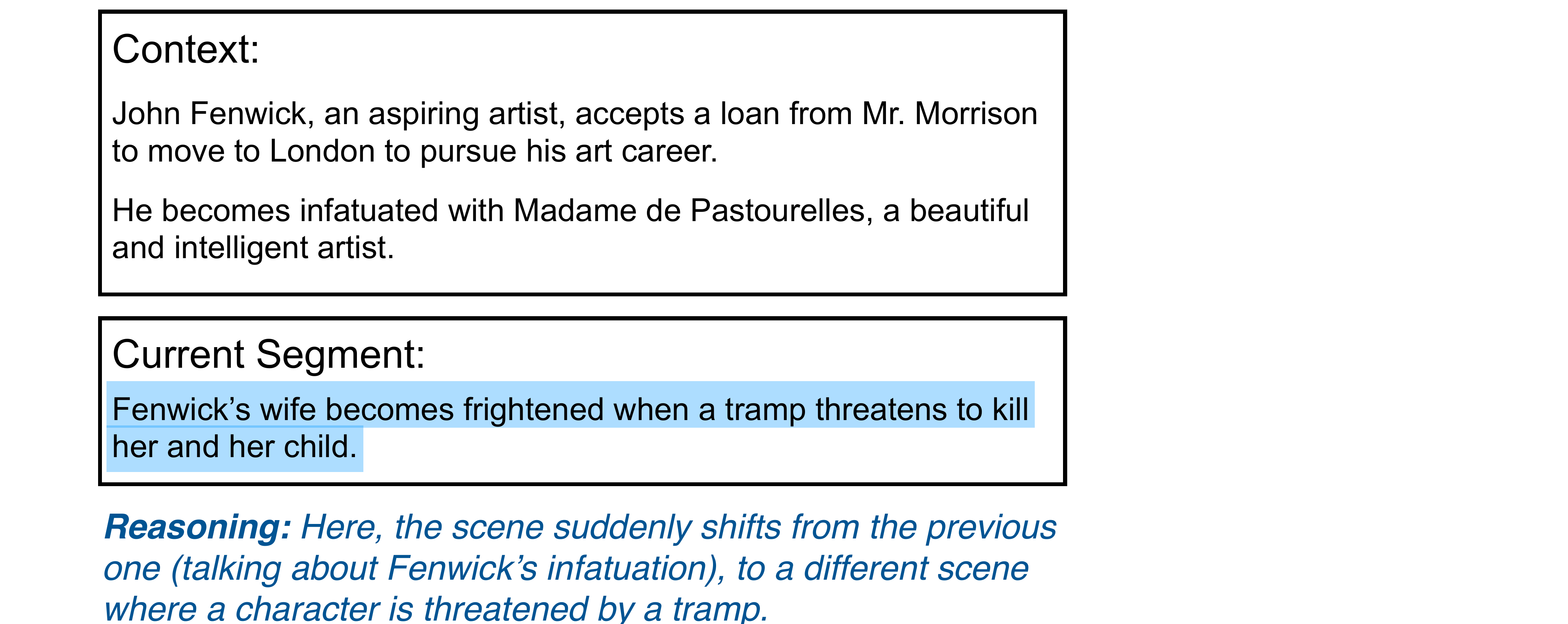}
    \caption{Illustration of \texttt{\textbf{SceneE}} errors provided to crowdworkers during training.}
    \label{fig:SceneE}
\end{figure}

These are called \textbf{Abrupt Transition from the Previous Scene
} in the task interface. We provide the illustrative example show in Figure \ref{fig:SceneE} along with the following definition:

\textit{``These refer to coherence errors where there is a sudden shift in the setting or the narrative in the story. These often happen in two scenarios:}
\begin{enumerate}[leftmargin=*]
    \item \textit{There is an abrupt change in the people/characters being discussed and/or an abrupt change in the surroundings/event.}
    \item \textit{Scenarios where the previous scene's phrasing strongly implies that more information/events are forthcoming, but the previous scene gets abruptly cut off and a completely new scene starts.}
\end{enumerate}

\textit{\textbf{Please choose full sentences as spans for this error type.}}''

\subsection{\texttt{\textbf{InconE}}}
\begin{figure}[h]
    \centering
    \includegraphics[trim=20mm 180mm 240mm 35mm, scale=0.25, clip]{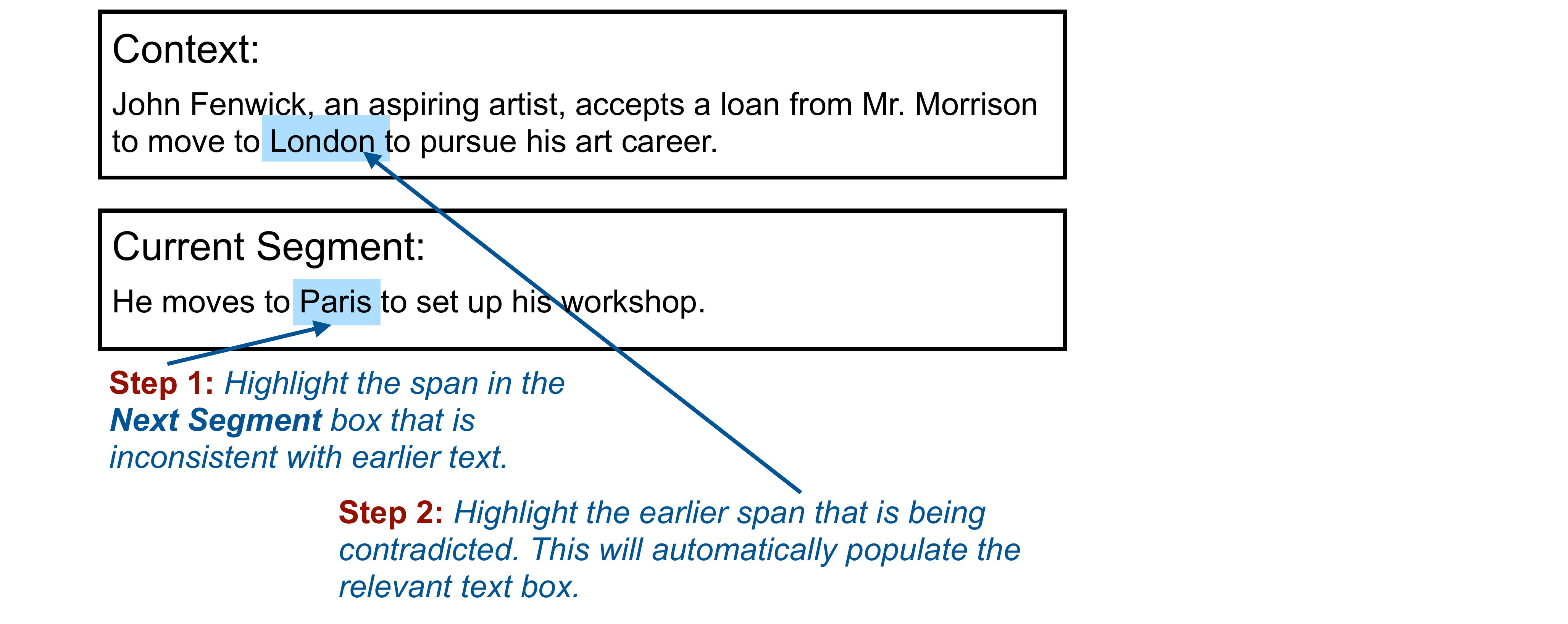}
    \caption{Illustration of \texttt{\textbf{InconE}} errors provided to crowdworkers during training.}
    \label{fig:inconE}
\end{figure}

Figure \ref{fig:inconE} shows an example of \textbf{Inconsistent} error shown to annotators.

\textit{``These refer to text spans that contradict previous content (either in the context or the next segment box itself.)}

\textit{Note: You will also be asked to highlight the `previous' span that is contradictory to the selected span. Highlighting this previous span (from either the context or the next segment box itself) will populate the relevant input box automatically.''}

\subsection{\texttt{\textbf{CorefE}}}
\begin{figure}[h]
    \centering
    \includegraphics[trim=20mm 250mm 240mm 35mm, scale=0.25, clip]{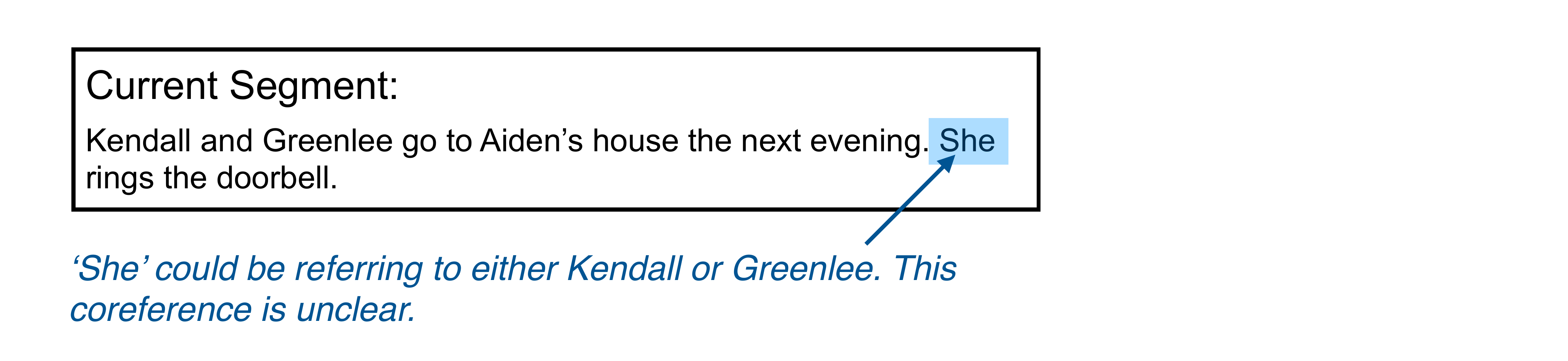}
    \caption{Illustration of \texttt{\textbf{CorefE}} errors provided to crowdworkers during training.}
    \label{fig:corefE}
\end{figure}

Figure~\ref{fig:corefE} shows an example of \textbf{Unclear Coreference} provided to annotators.

\textit{``These refer to errors where it is unclear who/what a pronoun or refers to.''}

\subsection{\texttt{\textbf{RepE}}}
\begin{figure}[h]
    \centering
    \includegraphics[trim=20mm 215mm 240mm 35mm, scale=0.25, clip]{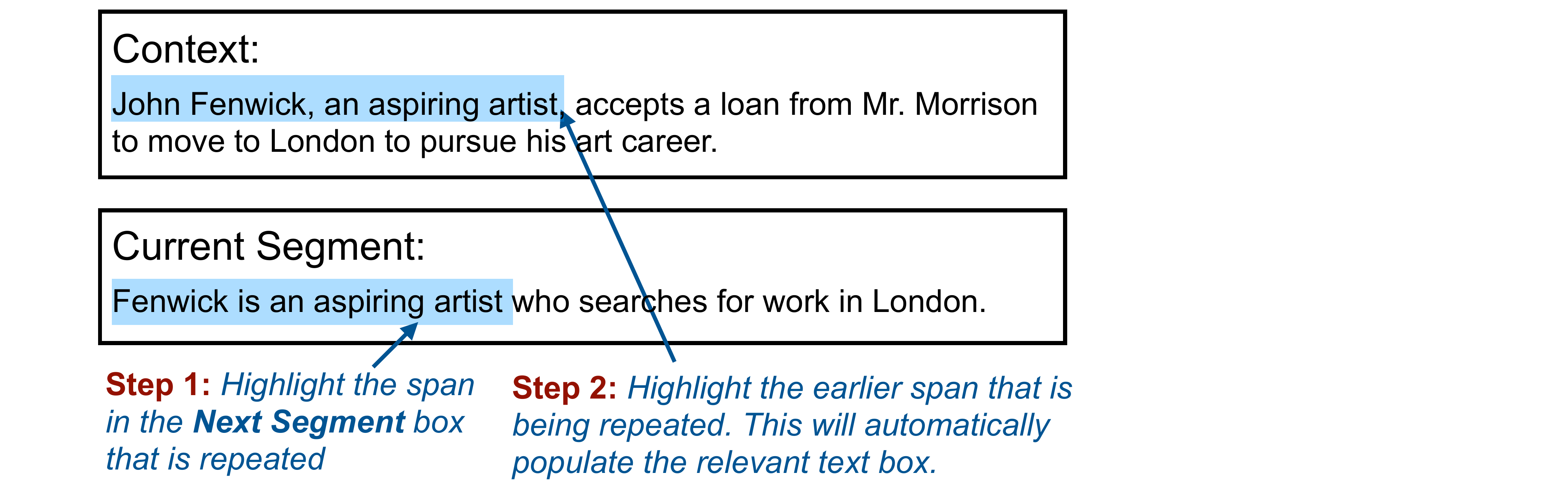}
    \caption{Illustration of \texttt{\textbf{RepE}} errors provided to crowd workers during training.}
    \label{fig:repE}
\end{figure}

Figure \ref{fig:repE} shows an example of \textbf{Repetition} errors.

``\textit{These refer to spans where content is repeated.}

\textit{Note: For these, you will also be asked to highlight the `previous' span that contains the same text/content as the selected span. Highlighting this previous span (from either the context or the next segment box itself) will populate the relevant input box automatically.}''

\subsection{\texttt{\textbf{GramE}}}
These are called \textbf{Ungrammatical/Nonsensical
} in the interface. 

``\textit{These refer to text spans that have grammar errors. Also included in this category are cases where there are obvious commonsense errors or the text does not make any sense at all.}''

\subsection{Task Interface}
\begin{figure}[t]
    \centering
    \includegraphics[trim=70mm 195mm 15mm 10mm,scale=0.29, clip]{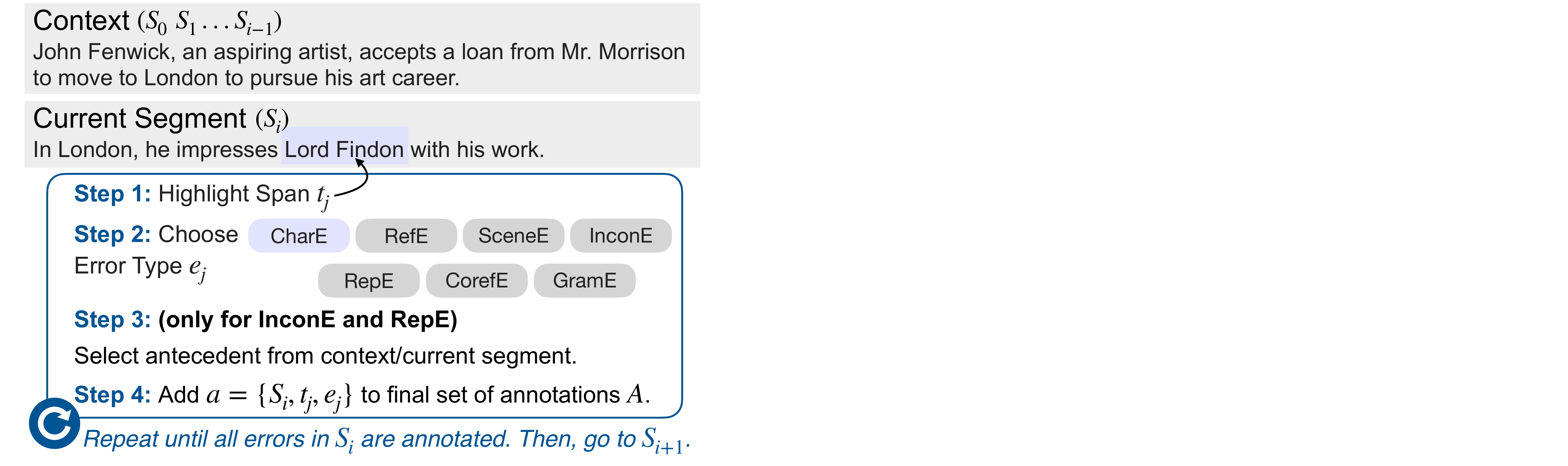}
    \caption{Workflow for annotating coherence errors in segment $S_i$ with respect to the context, i.e. $S_0, S_1, ..., S_{i-1}$.}
    \label{fig:task_interface}
\end{figure}

\begin{figure*}[h]
    \centering
    \includegraphics[width=\textwidth, clip]{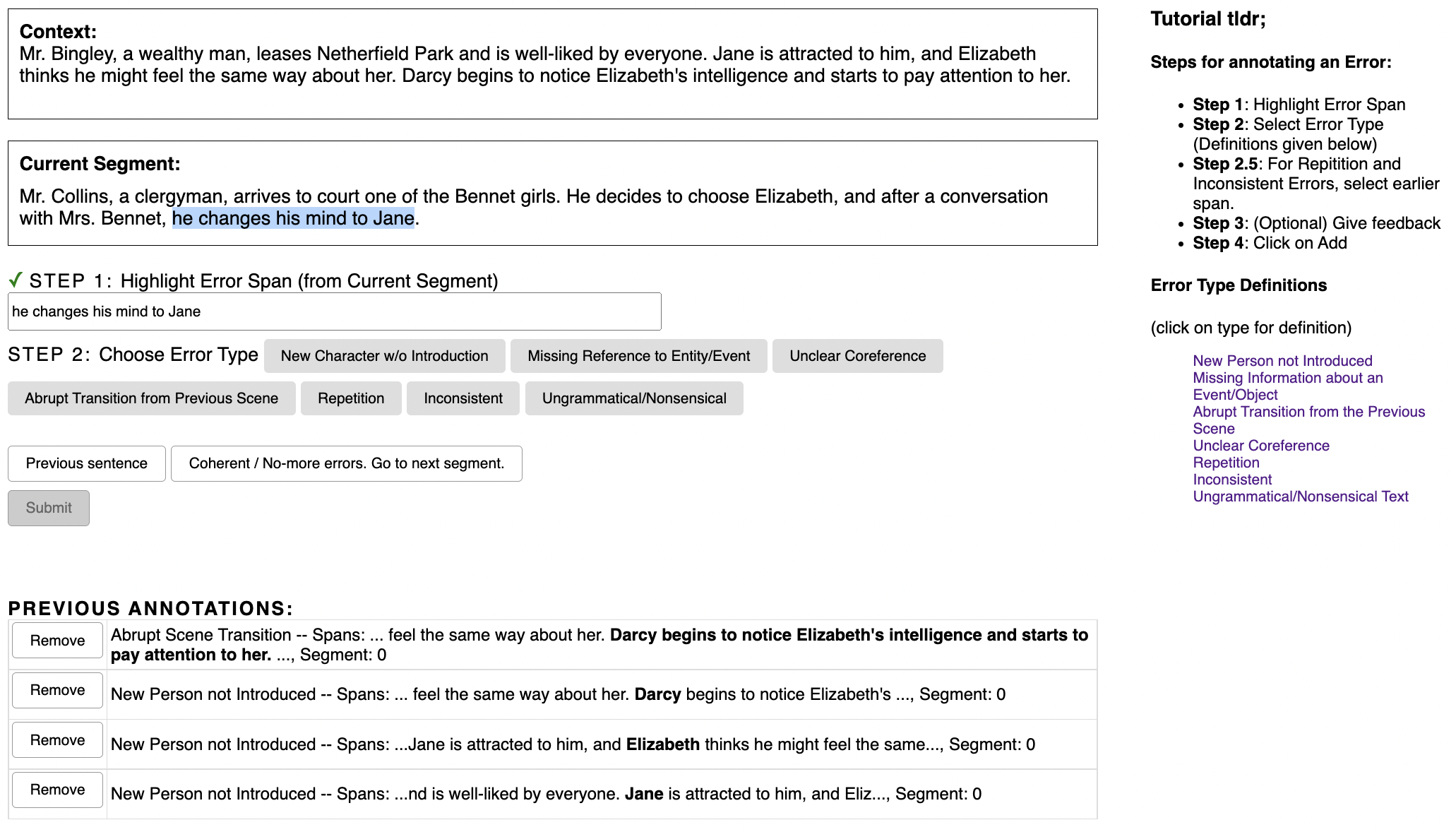}
    \caption{Screenshot of the task interface for \framename~annotations}
    \label{fig:task_interface_screenshot4}
\end{figure*}

In Section \ref{sec:workflow}, we described the annotation work for the \framename~framework. Figure \ref{fig:task_interface} visually illustrates this overall workflow for annotating errors in segment $S_i$. A screenshot of the actual task interface is shown in Figure \ref{fig:task_interface_screenshot4}.

\begin{figure*}[h]
    \centering
    \includegraphics[width=\textwidth, clip]{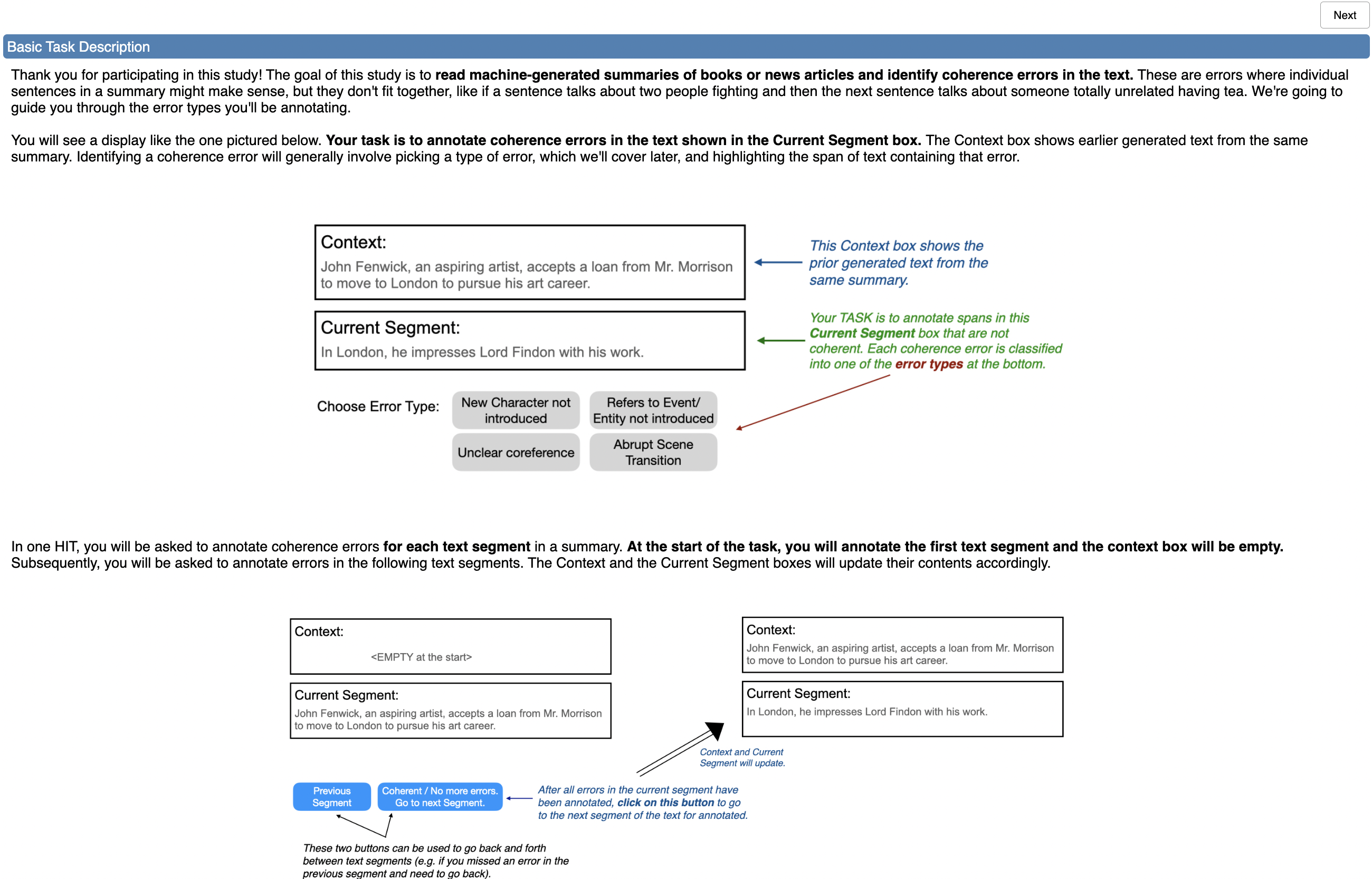}
    \caption{Screenshot of the first page of the tutorial provided to crowd annotators}
    \label{fig:task_interface_screenshot1}
\end{figure*}

\begin{figure*}[h]
    \centering
    \includegraphics[width=\textwidth, clip]{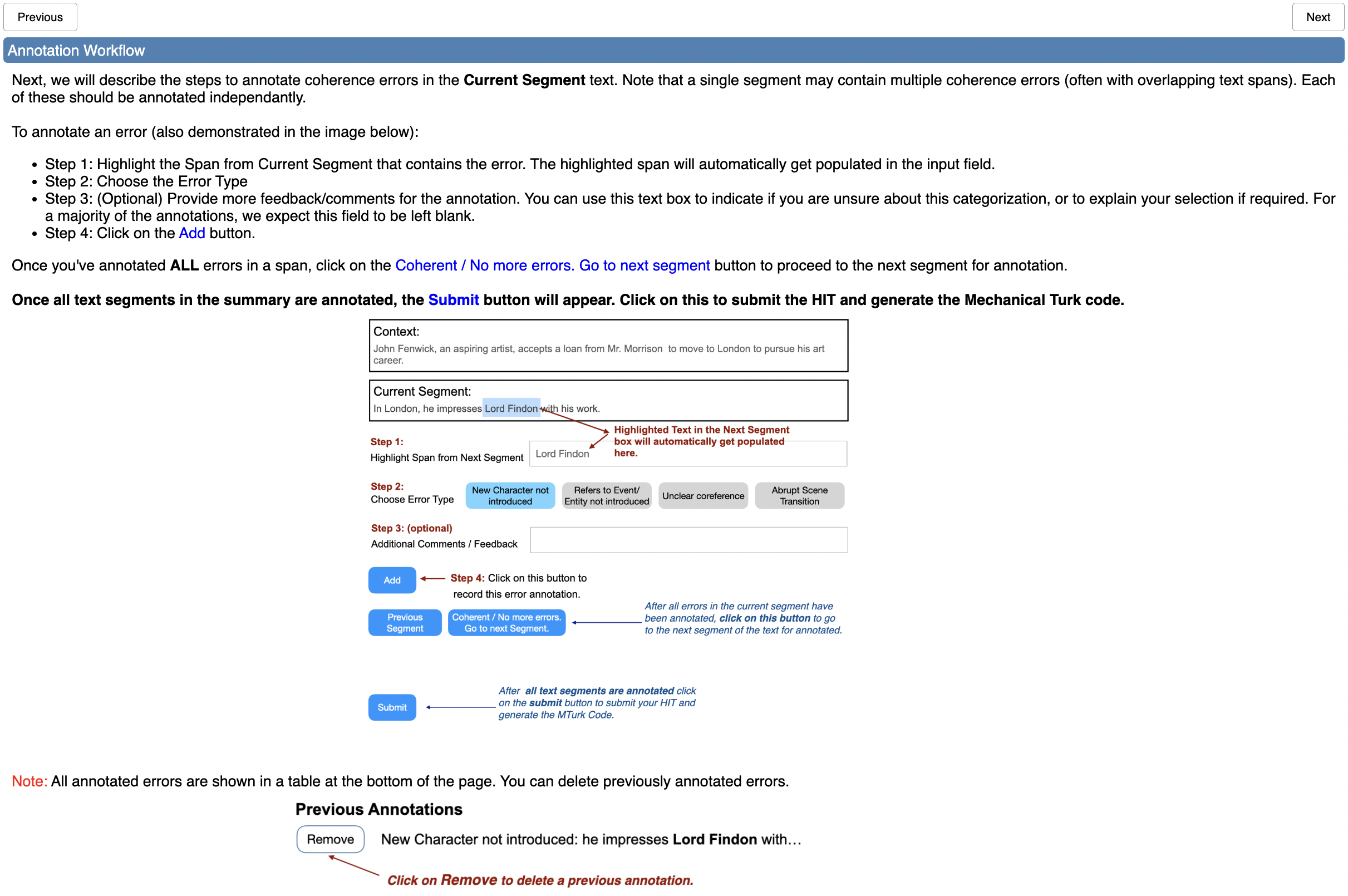}
    \caption{Screenshot of the second page of the tutorial provided to crowd annotators}
    \label{fig:task_interface_screenshot2}
\end{figure*}

\begin{figure*}[t]
    \centering
    \includegraphics[width=\textwidth, clip]{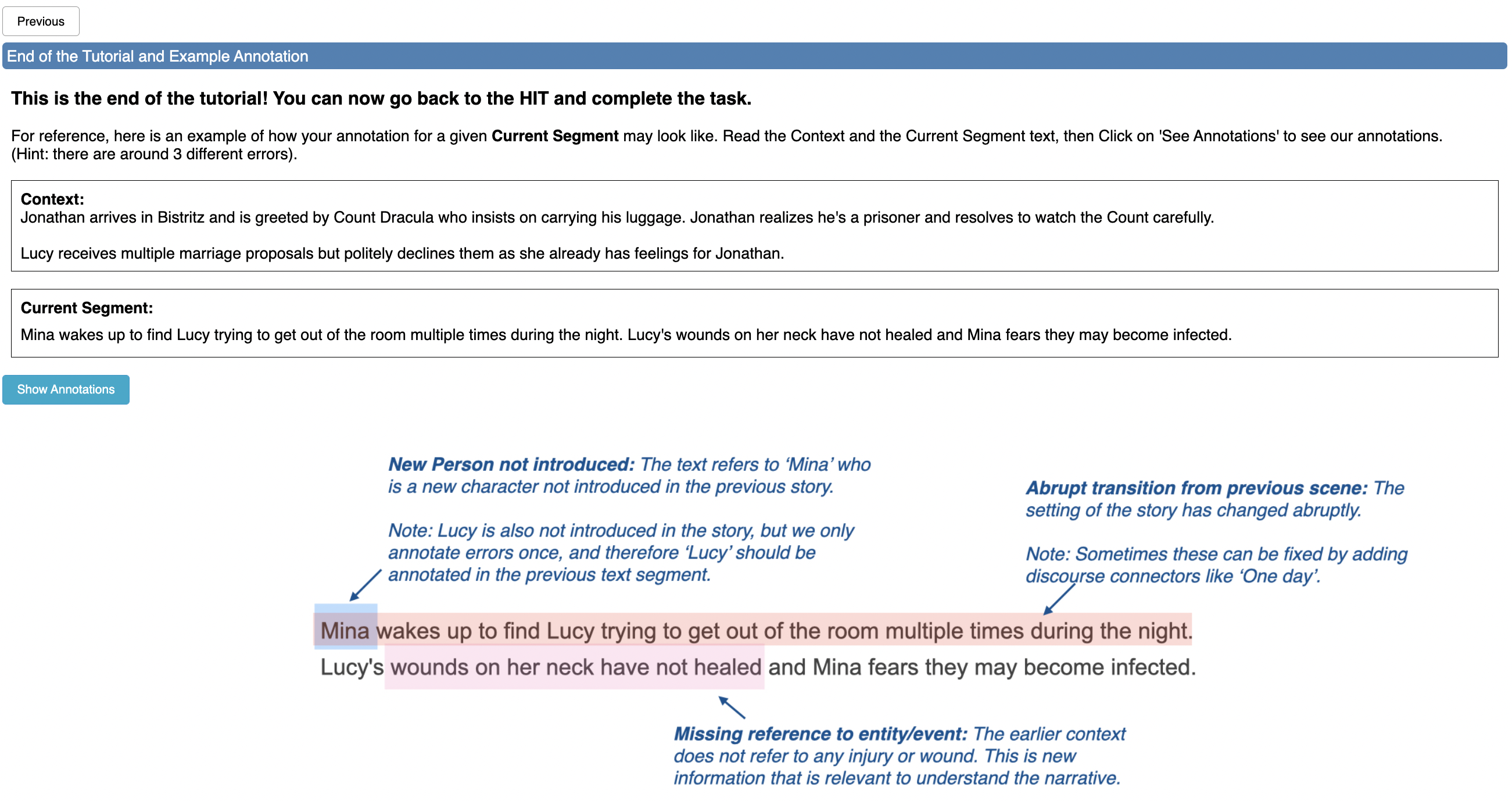}
    \caption{Screenshot of the last page of the tutorial provided to crowd annotators}
    \label{fig:task_interface_screenshot3}
\end{figure*}

We also include screenshots of our task instructions. Figure \ref{fig:task_interface_screenshot1} explains the basic task to the annotators. Figure \ref{fig:task_interface_screenshot2} shows the detailed task workflow and the steps to annotate errors in a text segment. Figure \ref{fig:task_interface_screenshot3} shows an example annotation with multiple coherence errors for reference.

\section{Datasheet for \framename}

\subsection{Motivation for Dataset Creation} 

\paragraph{Why was the dataset created?} Despite recent interest in long document summarization research and generation of long narrative summaries \cite{kryscinski2021booksum, zhang2021summ, mao2021dyle, wu2021recursively}, we lack evaluation frameworks to compare these approaches and measure progress. Current automatic and human evaluation methods fail to identify gaps in narrative coherence and are not suited for evaluating long summaries. Our \framename~dataset and annotation framework releases a large-scale dataset of fine-grained coherence annotations and establishes a protocol for eliciting such annotations from crowdworkers. This provides a foundation for future research efforts in this area.

\paragraph{Has the dataset been used already?} At the time of submission, the dataset has only been used in the current paper for analysis of generation errors made by current state-of-the-art summarization models and for training automatic coherence detection models. 

\paragraph{Who funded the dataset?}
We withhold this information to maintain anonymity but will include it upon publication.

\subsection{Dataset Composition}

\paragraph{What are the instances?} Each instance in this dataset is a model generated summary from either the book or the movie domain. All summaries are in the English language.

\paragraph{How many instances are there?} Our dataset contains annotations for 160 generated summaries (including both expert and crowd annotations). 

\paragraph{What data does each instance consist of?} Each instance contains multiple span-level highlights corresponding to coherence errors, each of which is tagged with a specific error category. 

\paragraph{Does the data rely on external sources?} Yes. For the book datasets,  we annotate summaries from the publicly available model outputs released by \citet{wu2021recursively}. For movies, we generate summaries using the Summ\^{}N model \cite{zhang2021summ} on the publicly available TRIPOD dataset \cite{papalampidi2020screenplay}.

\paragraph{Are there recommended data splits or evaluation measures?} We will include the recommended training, development, and test splits for our annotations with the dataset release. The statistics for the data splits are outlined in Section \ref{sec:benchmarking}.

\subsubsection{Data Collection Process}
\paragraph{Who was involved in the collection process and what were their roles?} For expert annotations, 3 authors of the paper with experience in engaging with model-generated text annotated 10 book summaries. To recruit crowd annotators, we launched a qualification task on Mechanical Turk. After this qualification, 11 workers were asked to annotate 150 summaries. 

\paragraph{How was the dataset collected?} 
Given a generated summary, annotators were asked to select span highlights that correspond with coherence errors and categorize the type of that error. We provided all annotators with detailed instructions describing the task interface, error type definitions as well as the overall workflow. 

\paragraph{Over what time frame was the data collected?} The dataset was collected over the months of March and April 2022.

\paragraph{Does the dataset contain all possible instances?} No, we only annotate narrative summaries from two summarization models on two domains (movies and books). Moreover, our dataset only contains English language summaries. 

\paragraph{If the dataset is a sample, then what is the population?} The dataset is a subset of generated summaries produced by state-of-the-art summarization models on narratives like books or movie screenplays. 

\subsection{Data Preprocessing}

\paragraph{What preprocessing/cleaning was done?} We fix sentence and word boundaries for highlighted spans from crowd annotations. 

\paragraph{Was the raw data saved in addition to the
cleaned data?} Yes.

\paragraph{Does this dataset collection/preprocessing procedure achieve the initial motivation?} Yes. This dataset serves as a large-scale collection of annotated coherence errors and provides the first characterization of such errors in long narrative summaries. 

\subsection{Dataset Distribution} 
\paragraph{How is the dataset distributed?} Our dataset is publicly released at this link: \url{https://github.com/tagoyal/snac}.

\paragraph{When was it released?} The dataset was released in October, 2022.

\paragraph{What license (if any) is it distributed under?} The dataset is released under the CC BY-SA 4.0 license.\footnote{\url{https://creativecommons.org/licenses/by-sa/4.0/legalcode}}

\paragraph{Who is supporting and maintaining the
dataset?} This dataset is maintained by authors of this paper. 

\subsection{Legal and Ethical Considerations}

\paragraph{Were workers told what the dataset would be
used for and did they consent?} Crowdworkers were aware that their responses were being collected as part of a research study on analyzing coherence errors in narrative text. The Amazon Mechanical Turk Participation Agreement permits the use of their annotated responses for this work. We do not release any personal information, e.g. worker IDs, of the crowdworkers.  

\paragraph{If it relates to people, could this dataset expose
people to harm or legal action?} No.

\paragraph{If it relates to people, does it unfairly advantage or disadvantage a particular social group?} No.

\end{document}